\documentclass[sigconf, nonacm]{acmart}
\AtBeginDocument{%
  }

\setcopyright{acmlicensed}
\copyrightyear{2026}
\acmYear{2026}
\acmDOI{XXXXXXX.XXXXXXX}
\acmConference[HRI '26]{Proceedings of the 2026 ACM/IEEE International Conference on Human-Robot Interaction}{March 16--19, 2026}{Edinburgh, Scotland, UK}

\usepackage{multirow}

\begin{document}

\title{WAFFLE: A Wearable Approach to Bite Timing Estimation in Robot-Assisted Feeding}

\author{Akhil Padmanabha}
\authornote{Equal Contribution.}
\affiliation{
  \institution{Carnegie Mellon University}
  \city{Pittsburgh}
  \state{PA}
  \country{USA}
}
\email{akhilpad@andrew.cmu.edu}

\author{Jessie Yuan}
\authornotemark[1] 
\affiliation{
  \institution{Carnegie Mellon University}
  \city{Pittsburgh}
  \state{PA}
  \country{USA}
}
\email{jzyuan@andrew.cmu.edu}

\author{Tanisha Mehta}
\affiliation{
  \institution{Carnegie Mellon University}
  \city{Pittsburgh}
  \state{PA}
  \country{USA}
}
\email{tsmehta@andrew.cmu.edu}

\author{Rajat Kumar Jenamani}
\affiliation{
  \institution{Cornell University}
  \city{Ithaca}
  \state{NY}
  \country{USA}
}
\email{rj277@cornell.edu}

\author{Eric Hu}
\affiliation{
  \institution{Cornell University}
  \city{Ithaca}
  \state{NY}
  \country{USA}
}
\email{elh245@cornell.edu}

\author{Victoria de León}
\affiliation{
  \institution{Carnegie Mellon University}
  \city{Pittsburgh}
  \state{PA}
  \country{USA}
}
\email{vrodrig2@andrew.cmu.edu}

\author{Anthony Wertz}
\affiliation{
  \institution{Carnegie Mellon University}
  \city{Pittsburgh}
  \state{PA}
  \country{USA}
}
\email{awertz@cs.cmu.edu}

\author{Janavi Gupta}
\affiliation{
  \institution{Carnegie Mellon University}
  \city{Pittsburgh}
  \state{PA}
  \country{USA}
}
\email{janavig@andrew.cmu.edu}

\author{Ben Dodson}
\affiliation{
  \institution{Cornell University}
  \city{Ithaca}
  \state{NY}
  \country{USA}
}
\email{bzd4@cornell.edu}

\author{Yunting Yan}
\affiliation{
  \institution{Cornell University}
  \city{Ithaca}
  \state{NY}
  \country{USA}
}
\email{yy2244@cornell.edu}

\author{Carmel Majidi}
\affiliation{
  \institution{Carnegie Mellon University}
  \city{Pittsburgh}
  \state{PA}
  \country{USA}
}
\email{cmajidi@andrew.cmu.edu }

\author{Tapomayukh Bhattacharjee}
\authornote{Equal Advising.}
\affiliation{
  \institution{Cornell University}
  \city{Ithaca}
  \state{NY}
  \country{USA}
}
\email{tapomayukh@cornell.edu}

\author{Zackory Erickson}
\authornotemark[2]
\affiliation{
  \institution{Carnegie Mellon University}
  \city{Pittsburgh}
  \state{PA}
  \country{USA}
}
\email{zackory@cmu.edu}

\renewcommand{\shortauthors}{Padmanabha\text{*}, Yuan\text{*} et al.}

\begin{abstract}
    \vspace{-0.15cm}
    Millions of people around the world need assistance with feeding. Robotic feeding systems offer the potential to enhance autonomy and quality of life for individuals with impairments and reduce caregiver workload. However, their widespread adoption has been limited by technical challenges such as estimating bite timing, the appropriate moment for the robot to transfer food to a user's mouth. In this work, we introduce \textbf{WAFFLE}: \textbf{W}earable \textbf{A}pproach \textbf{F}or \textbf{F}eeding with \textbf{LE}arned Bite Timing, a system that accurately predicts bite timing by leveraging wearable sensor data to be highly reactive to natural user cues such as head movements, chewing, and talking. We train a supervised regression model on bite timing data from 14 participants and incorporate a user-adjustable assertiveness threshold to convert predictions into proceed or stop commands. In a study with 15 participants without motor impairments with the Obi feeding robot, WAFFLE performs statistically on par with or better than baseline methods across measures of feeling of control, robot understanding, and workload, and is preferred by the majority of participants for both individual and social dining. We further demonstrate WAFFLE's generalizability in a study with 2 participants with motor impairments in their home environments using a Kinova 7DOF robot. Our findings support WAFFLE's effectiveness in enabling natural, reactive bite timing that generalizes across users, robot hardware, robot positioning, feeding trajectories, foods, and both individual and social dining contexts. Videos are located on our project website\footnote{\url{https://sites.google.com/view/bitetiming/}}\footnote{This research is supported by the National Science Foundation Graduate Research Fellowship Program under Grant No. DGE1745016 and DGE2140739.}.
    
\end{abstract}


\begin{teaserfigure}
  \vspace{-0.45cm}
  \centering
  \includegraphics[width=0.95\textwidth]{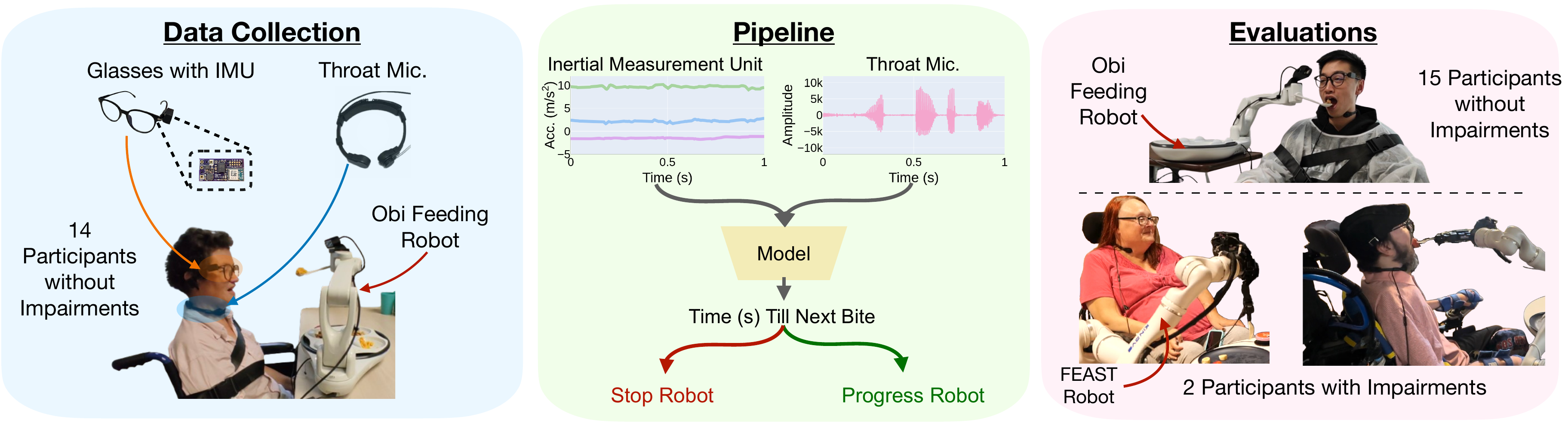}
 \vspace{-0.35cm}
 \caption{We present \textit{WAFFLE}, \textit{W}earable \textit{A}pproach \textit{F}or \textit{F}eeding with \textit{LE}arned Bite Timing, a bite timing algorithm for assistive feeding robots that generalizes across users with and without motor impairments, feeding robots, robot positioning, feeding trajectories, foods, and both individual and social dining settings. We collect data from 14 participants without impairments using the Obi robot to train a machine learning model that estimates the time until the next bite. At deployment, a user-selected assertiveness threshold is used to convert predictions into robot stop and progress commands. Lastly, we evaluate WAFFLE using two different robots with 15 participants without impairments and 2 participants with motor impairments in the home.}
 \vspace{-0.1cm}
  \label{fig:teaser}
\end{teaserfigure}


\maketitle

\vspace{-0.25cm}
\section{Introduction}
\vspace{-0.15cm}
According to the World Health Organization, nearly 190 million people worldwide live with motor impairments~\cite{taylor2018americans}, often needing assistance with Activities of Daily Living (ADLs) such as eating, bathing, and dressing~\cite{heimbuch2023prevalence, edemekong2019activities}. Specifically, around 3.5 million people in the United States alone require assistance with feeding, with this estimate increasing as the population ages~\cite{heimbuch2023prevalence}. For individuals with impairments, feeding assistance from a caregiver can yield feelings of anxiety and self-consciousness, and a sense of burden~\cite{jacobsson2000people, shune2020altered, nanavati2023design}. Physical caregiving robotics have shown their utility in a variety of activities of daily living (ADLs) including feeding assistance, increasing independence, quality of life, and well-being for individuals with impairments~\cite{madan2022sparcs, padmanabha2023hat, padmanabha2024independence, yang2023high, nanavati2023physically, nanavati2023design}. For feeding, although commercial robots like the FDA-compliant Obi Robot~\cite{Obi} are available, their adoption remains limited, in part due to technical limitations in food acquisition, bite transfer, bite timing, and user interfacing. Notably, Obi relies on manually programmed trajectories and does not adapt its motions to variations in user behavior or food positioning, limiting its flexibility in real-world use. Recent research has focused on addressing these challenges~\cite{jenamani2024feel, padmanabha2024voicepilot, sundaresan2023learning, ondras2022human, jenamani2025feast, nanavati2025lessons}. 

In this work, we focus on bite timing, the appropriate moment for a robot to deliver food to a user. Bite timing is an important component of robot-assisted feeding with prior research highlighting the value users place on having multiple options for triggering bite timing and on the adaptability of the timing itself~\cite{bhattacharjee2019community, nanavati2023design, jenamani2025feast}. Most existing approaches rely on users manually initiating bite timing, often through actions like opening their mouth. However, such methods can be distracting and intrusive, particularly in social dining contexts~\cite{ondras2022human, nanavati2023design}. While learning-based methods such as SoNNET~\cite{ondras2022human} aim to reduce disruptions and workload, they depend on constrained setups with fixed, external sensors and multiple diners, limiting practicality in diverse user settings. These limitations point to the need for approaches to bite timing that are both generalizable and practical across a wider range of real-world scenarios.

Caregivers often rely on behavioral cues, such as head movements, chewing, and speaking, to determine when to offer the next bite during assisted feeding. To sense and act on these cues, we present a system named \textbf{WAFFLE}: \textbf{W}earable \textbf{A}pproach \textbf{F}or \textbf{F}eeding with \textbf{LE}arned Bite Timing, consisting of wearable sensors for estimation of bite timing for both individual and social dining scenarios, designed to work with any robotic feeding platform positioning, any robot feeding trajectory, any foods, and any user. The wearables, an inertial measurement unit (IMU) mounted on a pair of glasses and a throat contact microphone worn around the neck, provide our system with cues often used by caregivers during the feeding process including a user's head movements and when they are chewing and talking. The wearables are privacy-preserving as the throat microphone does not record ambient sounds. 

As seen in Fig.~\ref{fig:teaser}, in order to develop a supervised learning algorithm for bite timing estimation, we conduct a data collection study with 14 individuals without motor impairments with the Obi robot~\cite{Obi} feeding from the front. We use this wearable data to train a bite timing regression algorithm to estimate the time until the next bite. To include user preference into how assertive the robot is during deployment, we use a simple user-selected threshold to convert predictions into progress / stop commands for the robot. In an initial evaluation study with 15 individuals without motor impairments using the Obi robot and feeding from the side, we compare our WAFFLE (Wearables) method with two baselines, Mouth Open and Fixed Interval. Results show that WAFFLE performs statistically on par with or better than the baselines in both individual and social dining across all metrics, including feeling of control, robot understanding, and seamlessness. We also find that the Wearables method is strongly preferred by participants in contrast to current bite timing approaches. Our method performs strongly in a second evaluation study with two participants with motor impairments and the FEAST system~\cite{jenamani2025feast}, which consists of a Kinova 7DOF robot arm feeding from the front with both in-mouth and out-of-mouth bite transfer, demonstrating its effectiveness and ability to generalize to participants with impairments and other feeding robots. Lastly, we find that participants are well accepting of our wearables, reporting that both the IMU-equipped glasses and throat microphone were unobtrusive and quickly faded from attention during use, further supporting the system's practicality for real-world  deployment in daily dining scenarios.


\vspace{-0.3cm}
\section{Related Work}
\vspace{-0.1cm}
\subsection{Robot-Assisted Feeding}
\vspace{-0.12cm}

Past research in robot-assisted feeding has focused primarily on food acquisition~\cite{gordon2020adaptive, sundaresan2023learning, gallenberger2019transfer, feng2019robot, sundaresan2022learning, liu2024adaptive, gordon2023towards, bhattacharjee2019towards, keely2024kiri, jenamani2024flair}, bite transfer~\cite{jenamani2024feel, belkhale2022balancing, shaikewitz2023mouth}, and system/interface design~\cite{gordon2024adaptable,padmanabha2024voicepilot, yuan2024towards, park2020active, nanavati2023design, bhattacharjee2019community, bhattacharjee2020more, barrue2024nyam, higa2014vision, jenamani2025feast, nanavati2025lessons}. For bite timing, many of these works have relied on manual triggering by participants through a button or assistive interface, fixed-interval timing, or open-mouth detection~\cite{nanavati2023design, perera2017eeg, schroer2015autonomous, park2020active, bhattacharjee2020more, jenamani2025feast}.

In contrast, Ondras et al. introduced the SoNNET algorithm, which employs three RGB-D cameras and a microphone array placed centrally on the dining table to predict bite timing during social dining scenarios~\cite{ondras2022human}. The authors collected an initial dataset with participants without impairments dining alongside two co-diners, capturing multimodal data under low background noise conditions. Their algorithm performs binary classification using features extracted from the RGB-D cameras and microphone array to determine whether the user intends to take a bite and triggers the robot to proceed along the entire feeding trajectory when a positive prediction is made. While their study found that participants preferred the predicted bite timing over two baseline methods (Mouth Open and Fixed Interval), no statistical significance was observed between SoNNET and the Mouth Open baseline for most metrics, indicating room for improvement. Furthermore, the system’s strong dependence on fixed sensor placement, presence of two co-diners, and controlled acoustic conditions limits its generalizability to everyday environments such as restaurants, where sensor occlusion, user diversity, and privacy concerns are more pronounced. The approach has also not been extended to individual dining or evaluated with individuals with motor impairments, different robot platforms, or varied robot and sensor orientations. In comparison, our method works across both individual and social dining scenarios, uses common wearable sensors, is robust to background acoustic noise, and has been evaluated with 2 participants with impairments and two different robotic systems in varied orientations, achieving statistically significant improvements across many metrics compared to the baseline methods.

\vspace{-0.3cm}
\subsection{Wearable Sensing}
\vspace{-0.05cm}
Wearable sensing involving accelerometers, inertial measurement units (IMUs), and contact microphones has been useful for identification of various human movements~\cite{heikenfeld2018wearable, yetisen2018wearables, padmanabha2023multimodal} and for control of mobile robots for activities of daily living during robotic caregiving tasks~\cite{padmanabha2023hat, padmanabha2024independence, padmanabha2025towards}. Past work has shown that wearables can successfully capture chewing and swallowing using acoustic, inertial, electromyography (EMG), piezoelectric, or a fusion of the aforementioned sensing methods~\cite{vu2017wearable, sazonov2008non, bedri2020fitbyte, prioleau2017unobtrusive}. Positioning of the sensors on the human body is important with wearable sensors often mounted on the inner/outer ear, head, neck, and chest~\cite{vu2017wearable}. Visual approaches using external mounted cameras have shown promise for chewing and swallowing detection but require the user to be within the view of the camera continuously~\cite{tufano2022capturing, hossain2020automatic}. Insights from these past works on the types of wearable sensing methods and the placements of these sensors show promise for using a fusion of wearable sensing modalities, specifically IMUs and contact microphones, for estimation of bite timing for robot-assisted feeding.   

\vspace{-0.3cm}
\section{Methods}
\vspace{-0.05cm}
Bite timing is a challenging problem: it often depends on a subconscious sense of when it feels appropriate to take a bite of food, especially during social dining scenarios. The problem becomes more complicated since there are often many valid timings and robot motions that result in an appropriate bite. Bite timing is also  dynamic; for example, a person may lift food to their mouth, pause, and only proceed to eat at a very specific moment. We hypothesize that a robot can achieve appropriate bite timing by being highly reactive to natural user cues from wearable sensor data, including chewing, speaking, and head movements. Our goal is to develop a system that is generalizable across all variables in a robot-assisted feeding setting, including different users, feeding robots, robot positioning, feeding trajectories, foods, and individual and social dining scenarios. To use wearable data to estimate bite timing, we propose a supervised data-driven algorithm. In the following sections, we describe our methodology for data collection, the architecture of our learned model, and our deployment approach, followed by a description of two evaluation studies involving both individuals with and without impairments. The setup for all three studies are shown in Fig.~\ref{fig:study_3_phases} and compared in table format in Table~\ref{tab:study_comparison}.

\begin{figure*}[t!]
  \vspace{-0.5cm}
  \centering
  \includegraphics[width = 0.9\textwidth]{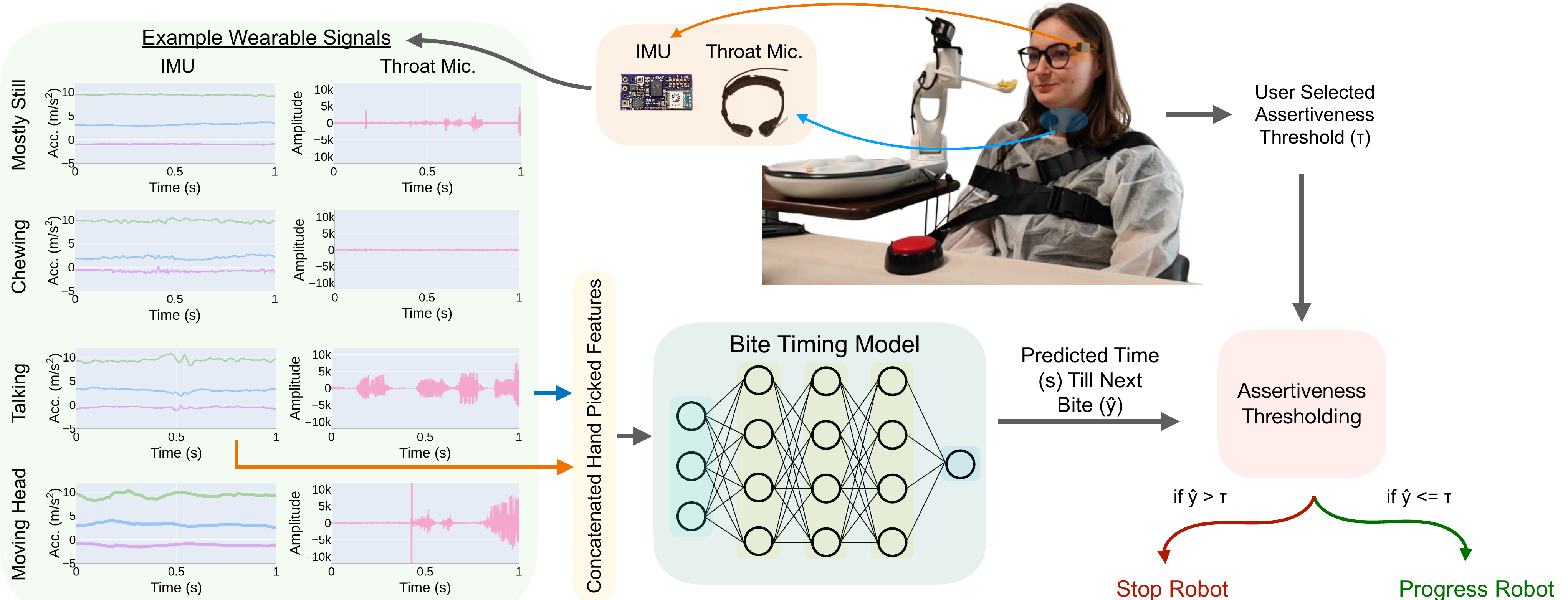}
  \vspace{-0.3cm}
  \caption{WAFFLE Deployment Pipeline. Features from 1 second windows of IMU and Throat Microphone data are inputted into our bite timing machine learning model which outputs $\hat{y}$, the predicted time (s) until the next bite. We use a user selected assertiveness threshold, $\tau$, to stop and progress the robot along the trajectory.}
  \Description{TODO}
  \vspace{-0.5cm}
  \label{fig:model_pipeline}
\end{figure*}

\vspace{-0.25cm}
\subsection{Phase 1: Data Collection Study}
\vspace{-0.05cm}
To collect a dataset of wearable signals for training a supervised learning algorithm, we conducted a human study, referred to as Phase 1, with 14 participants (7 M, 7 F) ranging in age from 19 to 60 (Mean = 26.07, SD = 10.16). We received approval from Carnegie Mellon University's Institutional Review Board. Demographic data is presented in Appendix~\ref{appendix:phase1_demographics}.

As illustrated in Fig.~\ref{fig:study_3_phases}, participants wear two wireless Inertial Measurement Units (IMUs) mounted on glasses, two additional wireless IMUs embedded in earbuds, and a commercially available throat microphone. The IMU boards~\cite{SMLPosey2025}, shown in Fig.~\ref{fig:model_pipeline}, are custom designed and fabricated and are housed in 3D-printed enclosures affixed to the glasses and in the 3D-printed earbuds. Participants sit in a wheelchair with their movement restricted by two straps, simulating the condition of someone with severe motor impairments as suggested by an Occupational Therapist consultant. The Obi robot is situated in front of the participant with an Intel Realsense D435i RGBD camera mounted on the robot arm. A companion laptop running ROS 1 Noetic is used to simultaneously collect RGB images and depth images at 15 Hz, 3-axis accelerations and quaternions from each of the four IMUs at 200 Hz, and audio from the throat microphone at 44.1 kHz. Two push buttons are additionally connected to the laptop and are used to control the Obi robot. Before starting data collection, we kinesthetically move the robot to record the feeding position directly in front of the user's mouth. Next, we introduce the participant to the system by having them eat 2 bites of food and to try rotating between the 4 bowls using the red button, seen in Fig.~\ref{fig:model_pipeline}. Information on the foods utilized is included in Appendix~\ref{appendix:foods}. 

We structure our data collection as follows. The robot automatically scoops food from the selected bowl and moves to a predefined staging point about 15 inches from the participant's mouth. Six participants do a participant-controlled version of the study, while the other eight participate in a Wizard of Oz (WoZ) version. For the participant-controlled version, between the staging and feeding points, the participants use a yellow button, which moves the robot towards the mouth when it is pushed and held down. When the button is released, the robot doesn't move. For the Wizard of Oz version, participants are told that the robot is fully autonomous, and a researcher uses the yellow button to control the robot between the staging and feeding points. The researcher specifically looks for cues from the user that caregivers often use, including chewing, talking, and glancing at the robot. They follow a set of guidelines we developed, provided in Appendix~\ref{appendix:wizardofoz}, to decide when to progress the robot and stop it.

We conduct both the Wizard of Oz and participant-controlled versions of the study because each provides valuable bite timing data for training our algorithms. The Wizard of Oz condition offers more structured training data as the experimenter directly controls the robot’s movements, ensuring consistent robot movements across participants. This structure makes it easier for the model to learn generalized patterns by minimizing variability. Additionally, the Wizard of Oz condition enables us to collect data on interactions with an `autonomous' robot, including participants’ reactions to occasional bite timing errors. However, structured data alone may not capture the full range of nuanced user preferences in robot motion and bite timing, which is why participant-controlled data is additionally collected. In both conditions, the researcher initiates the next scooping action through a manual command entered on the companion laptop once the participant has completed the bite.

For both participant-controlled and WoZ studies, for every participant, we conduct both individual and social dining scenarios, with the order counterbalanced for all participants. For individual dining, the user has 6 minutes to eat using the robot without any external interactions. For social dining, the user has 12 minutes to eat using the robot while conversing with a researcher who sits in front of them. More details on the social dining setup are located in Appendix~\ref{appendix:socialdining}. During the debriefing of the Wizard of Oz version of the study, none of the participants reported suspicion that the robot was being controlled by the researcher.

\vspace{-0.25cm}
\subsection{Bite Timing Algorithm}
\label{sec:bite_timing_model}
\vspace{-0.1cm}

We introduce a regression model that predicts the number of seconds until the next bite using data from a throat microphone and a single IMU mounted on the left side of a glasses frame. As shown in the raw data signals in Fig.~\ref{fig:model_pipeline}, the IMU's acceleration signals and the throat microphone's signals offer complementary insights into head movements, chewing, and talking. The IMU effectively captures head movements and chewing activity, while the throat microphone excels at detecting vibrations from swallowing and vocal cord activity during speech. To ensure the approach generalizes across different robot platforms, user orientations, and feeding trajectories, we intentionally exclude RGB camera data from the Phase 1 study. In real-world settings, feeding robot end effectors with cameras are often positioned above the plate or away from the user to avoid obstructing social interactions. Even when oriented toward the user, natural head movements can shift the face out of view, and because the camera is mounted on the robot arm, facial keypoints are heavily influenced by the robot’s own movement rather than the participant’s behavior. As a result, camera-based features are unreliable for capturing user intent. We also omit the remaining three IMUs (right glasses, left/right earbud), as their signals are highly similar to that of the left glasses IMU.

Our bite timing algorithm is visually shown in Fig.~\ref{fig:model_pipeline}. For every 1 second (s) window of 3-axis IMU accelerations and throat microphone signal, we divide this into two 500ms windows and extract hand-picked features. Using low-level windows with hand-picked features has shown promise for activity recognition including chewing detection in past works~\cite{padmanabha2025egocharm, rosen2022charm, bedri2020fitbyte}. These features are concatenated and inputted into a multi-layer perceptron. The final layer directly outputs the predicted time (s), $\hat{y}$, until the next bite. The model is trained using ground truth labels derived from the Phase 1 dataset, where each label represents the time from a given sample to the timestamp at which the robot reaches the set feeding position in front of the user’s mouth. We evaluate using leave one subject out cross validation (LOSO-CV) and train the final model for evaluation studies on data from all 14 participants. We provide additional details on our bite timing model architecture in Appendix~\ref{appendix:modeling}.

To deploy our trained model and enable reactive robot behavior, we use a thresholding approach to convert the predicted time until the next bite into a binary proceed/stop command for the robot. If the predicted time ($\hat{y}$) until the next bite is such that $\hat{y} > \tau$, a stop moving command is sent to the robot. Otherwise, if $\hat{y} <= \tau$, then a proceed moving command is sent to the robot. We define the threshold value, $\tau$, as the user-selected assertiveness threshold, which determines how confidently the robot acts on its predictions. Lower values make the robot behave more cautiously, while higher values result in more assertive feeding actions. This threshold incorporates user preferences and interpretability, inspired by observations from our Phase 1 study where participants exhibited different preferences for how assertively they fed themselves during the participant-controlled version. 

To quantify how well our bite timing model would perform in deployment with thresholding, we use the ground truth 0 and 1 robot motion labels from the Phase 1 study, which were provided by the participant for the participant-controlled studies and by the researcher for the Wizard of Oz studies. We generate predictions (in seconds) from our model and convert these to binary (0 and 1) predictions using the threshold. We then calculate binary metrics using LOSO-CV, evaluating only the participants who completed the participant-controlled Phase 1 study. The binary classification metrics we use are accuracy and the normalized Matthews Correlation Coefficient (nMCC), where 0 indicates perfect negative correlation, 0.5 represents no correlation, and 1 indicates perfect positive correlation. We chose nMCC because it captures the quality of both true positive and true negative predictions which are equally important in our context, as they represent the robot's decisions to reactively proceed or stop along the feeding trajectory. These binary metrics do not reflect the appropriateness of the bite timing itself, but instead serve as an indicator of how closely the robot’s start and stop actions (as determined by thresholding) align with the executed trajectories from the Phase 1 study.

We calculate the aforementioned binary metrics using a fixed threshold of $\tau = 6$ seconds which produced the highest nMCC and accuracy scores across all discrete threshold values from 4 to 8 seconds. These results are provided in Appendix~\ref{appendix:fixed_thresholds}. We also found a fixed threshold of $\tau = 6$ seconds worked best in practice through testing among the research team and pilot studies. To show the benefit of user-selected variable thresholds, we also calculate binary metrics using the optimal threshold for each participant, out of $\tau \in \{4, 5, 6, 7, 8\}$ seconds, that leads to the highest nMCC score. In a deployment, we assume that participants would be able to choose this optimal threshold for themselves after getting fully acclimated with the system. During deployment, we also transform the assertiveness threshold (originally in units of seconds) into a mapped assertiveness threshold. This new scale ranges from 1 (least assertive) to 5 (most assertive).

\begin{table*}[h!]
\vspace{-0.1cm}
\centering
\caption{Accuracy and nMCC after Thresholding using Optimal Thresholds for each Participant and a Fixed Threshold}
\vspace{-0.4cm}
\begin{tabular}{|l|c|c|c|c|c|}
\hline
 & \textbf{IMU \& Throat Mic.} & \textbf{IMU only} & \textbf{Throat Mic. only} & \textbf{Always Feed} \\
 \hline
 \hline
\textbf{Accuracy (Optimal Thresholds)} & \textbf{0.700 $\pm$ 0.090} & 0.680 $\pm$ 0.071 & 0.532 $\pm$ 0.137 & 0.396 $\pm$ 0.122 \\ \hline
\textbf{Accuracy (Fixed Threshold)} & 0.668 $\pm$ 0.066 & 0.633 $\pm$ 0.069 & 0.523 $\pm$ 0.121 & 0.396 $\pm$ 0.122 \\ \hline \hline
\textbf{nMCC (Optimal Thresholds)} & \textbf{0.674 $\pm$ 0.088} & 0.655 $\pm$ 0.053 & 0.576 $\pm$ 0.065 & 0.500 $\pm$ 0.000 \\ \hline
\textbf{nMCC (Fixed Threshold)} & 0.650 $\pm$ 0.082 & 0.624 $\pm$ 0.065 & 0.554 $\pm$ 0.102 & 0.500 $\pm$ 0.000 \\ \hline
\end{tabular}%
\label{tab:thresholding_metrics}
\end{table*}

\begin{figure*}[ht!]
  \vspace{-0.25cm}
  \centering
  \includegraphics[width = 0.9\textwidth]{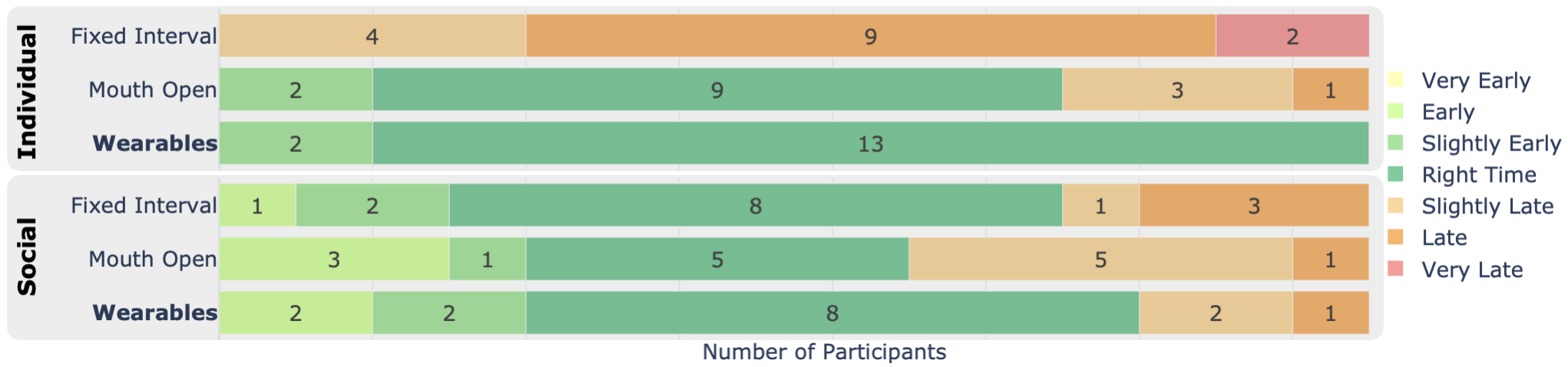}
  \vspace{-0.4cm}
  \caption{Phase 2 participant responses to the question, ``How was the bite timing of this method on average?'' for both individual and social dining scenarios. Data points are additionally provided in Table~\ref{tab:bite_timing}.}
  \Description{TODO}
  \label{fig:bite_timing}
  \vspace{-0.5cm}
\end{figure*}

\vspace{-0.3cm}
\subsection{Phase 2: Evaluation with Individuals without Impairments}
\vspace{-0.1cm}
\label{sec:phase2_methods}
In order to directly compare our wearables method with bite timing baselines, we conducted a comparison study, named Phase 2, with individuals without impairments. This study was completed with 15 participants (6 M, 9 F) with ages ranging from 19-28 (Mean = 22.27, SD = 3.1). We received approval from Carnegie Mellon University's Institutional Review Board. All demographic data is presented in Appendix~\ref{appendix:phase2_demographics}.


The two baselines we used were Fixed Interval and Mouth Open. In line with past work on bite timing~\cite{ondras2022human}, for the Fixed Interval method, the robot feeds the participant every 45 seconds. The Mouth Open method is the most commonly used trigger for bite timing in robot-assisted feeding studies~\cite{nanavati2023design, perera2017eeg, schroer2015autonomous, park2020active, bhattacharjee2020more}. For the Mouth Open baseline, after reaching the staging point, the robot announces ``Please open your mouth when you are ready to eat.'' The participant must then turn towards the Intel Realsense D435i RGBD camera mounted on the robot's arm and open their mouth to indicate they are ready for the next bite. Once the robot detects the user's mouth is open, it proceeds automatically through the entire feeding trajectory. 

Similar to the Phase 1 study, participants sit in a wheelchair with their movement restricted by two straps, simulating the condition of someone with severe motor impairments. Participants are outfitted with only a single wireless Inertial Measurement Units (IMU) mounted on the left side of the glasses and the throat microphone. Additionally, we place the commercially available Obi assistive feeding robot from DESĪN LLC~\cite{Obi} to the side of the participant, to highlight the ability of our algorithms to generalize to any robot positioning and feeding trajectory. Additional details on the foods used is included in the Appendix~\ref{appendix:foods}. Similar to Phase 1, we kinesthetically teach the robot the feeding position.

We begin with a practice session to familiarize participants with the system and the three feeding methods. For the two baseline methods, we explain their mechanisms and allow participants to try each once. For the wearables method, participants complete three practice trials to experience how specific cues including speaking, chewing, and large head movements, pause the robot’s motion, while minimal head movement allows it to continue along its trajectory. To minimize participant confusion during the study and due to the limited number of trials per method, we fix the mapped assertiveness threshold at a value of 3, which based on quantitative results presented in Section~\ref{sec:model_performance} and piloting, offers a balanced level of performance for most participants.

There are two parts to this study: individual dining and social dining. We counterbalance the order of these parts. For each part, we conduct three consecutive trials for each method. The methods and their relative order are blinded to the participant and counterbalanced across participants. The relative order of the methods stays consistent for both individual and social dining scenarios for each participant to reduce confusion and minimize cognitive load. The social dining scenario is conducted with a researcher as detailed in Appendix~\ref{appendix:socialdining}. After each method, the participant responds to questions related to bite timing and multiple 7-point Likert items related to distraction, feeling of control, robot understanding, appropriateness of robot movement, seamlessness, mental workload, physical workload, and natural conversation. After each part of the two parts of the study, the participant blindly ranks the three methods based on their preference. At the end of the study, the methods are unblinded and the participant ranks the three methods overall. Finally, they answer qualitative questions. All questions and Likert items are included in Appendix~\ref{appendix:study_setup}.

\vspace{-0.25cm}
\subsection{Phase 3: Evaluation 
with Individuals with Motor Impairments}
\label{sec:phase3_methods}
\vspace{-0.1cm}
We conducted a final study, named Phase 3, with 2 individuals with motor impairments in their homes. The two participants are referred to henceforth as I1 and I2. We received approval from Cornell University's Institutional Review Board. I1 is a 47-year-old male with a C4–C6 Spinal Cord Injury, with limited head and neck mobility, requiring inside-mouth bite transfer. I2 is a 48-year-old female with Multiple Sclerosis who prefers out of mouth bite transfer. I1 is fed banana, strawberry, and cantaloupe pieces while I2 is fed banana and strawberry pieces.  We use the FEAST system, which consists of a Kinova 7DOF arm, with the same safety features as described in Jenamani et al.~\cite{jenamani2025feast}. Both participants have experience with the FEAST system with manual triggering for bite timing in a past research study; I1 previously used a web interface and nodding and I2 previously used the web interface~\cite{jenamani2025feast}. To control the computer cursor for the web interface, I1 used a mouse tracker that follows a reflective dot on his nose while I2, with limited limb mobility, used her left arm.

For this study, we only evaluate our Wearables method. The FEAST robot automatically acquires the pieces of fruit and moves to a staging point. From there, the robot receives stop and proceed commands based on assertiveness thresholding on the predictions from our bite timing model. We set a fixed outside-mouth bite transfer distance, and the FEAST robot autonomously perceives the participant's mouth location and moves to that position. We make a minor modification to the Wearables method by having the robot commit to completing the remainder of the trajectory if it is within 5 centimeters of the participant's mouth. This change was motivated by observations in our Phase 2 study, where participants would open their mouths in anticipation of the robot finishing the trajectory, inadvertently causing it to stop. 

Similar to the Phase 2 study, we have a practice session for the Wearables method. During this session, we allow participants to test the various user selected assertiveness thresholds. I1 only tried mapped assertiveness level of 1 (least assertive) and used this level throughout the study. Meanwhile I2 tried 1, 3, and 5 and used 5 (most assertive) throughout the study. We counterbalance the order of the individual and the social dining sessions for the two participants. For each of the sessions, the participants eat six consecutive bites. Both participants answered similar quantitative and qualitative questions to the Phase 2 Study, included in Appendix~\ref{appendix:phase3_questions_responses}.


\vspace{-0.3cm}
\section{Results and Discussion}
\vspace{-0.1cm}
All participant demographics, quantitative and qualitative responses, and statistical testing results are included in the Appendix.

\vspace{-0.3cm}
\subsection{Model Performance}
\label{sec:model_performance}
\vspace{-0.1cm}

 As detailed in Section~\ref{sec:bite_timing_model}, our bite timing regression model outputs predictions of time until the next bite in seconds. Using LOSO-CV on the Phase 1 data, the combined IMU and throat microphone model achieves the lowest MAE of $2.83 \pm 0.33$ s. The IMU-only model performs slightly worse at $2.98 \pm 0.25$ s, while the throat microphone–only model reaches $3.16 \pm 0.27$ s. In comparison, a naive predictor using the mean of the training labels for each fold yields $3.25 \pm 0.15$ s. These metrics show that our model is able to learn a general relationship between 1-second wearable data windows consisting of motion, chewing, and talking signals and the time until the next bite, and they also highlight the value of combining sensor modalities over using a single sensing modality.
 
As described in Section~\ref{sec:bite_timing_model}, in order to deploy our method and enable reactive robot behavior based on wearable data, we threshold predictions from our bite-timing model. We calculate binary metrics (nMCC and accuracy) using both a fixed assertiveness threshold, $\tau = 6$ s, equivalent to a mapped assertiveness threshold of 3, and the optimal threshold per user. As previously mentioned, bite timing is a challenging problem and is largely subconscious, with multiple timings and robot motions that participants would accept. These binary metrics should not be interpreted as an indication of whether or not the bite timing was appropriate and instead be used to provide a signal on how well the robot's start and stop motions after thresholding align with the executed trajectories in the Phase 1 study.

As seen in Table~\ref{tab:thresholding_metrics}, our IMU \& Throat Mic. method achieves the best alignment performance of 0.700 accuracy and 0.674 nMCC using the optimal threshold per participant. This performance slightly outperforms the IMU only method which achieves 0.68 accuracy and 0.655 nMCC. We see significantly better alignment performance than an Always Feed method which achieves 0.3957 accuracy and 0.5 nMCC. We also find that our combined IMU \& Throat Mic. algorithm outperforms both the IMU only and Throat Mic. only algorithms using these binary metrics, further backing our proposed approach of using the combination of the two modalities.

Our metrics additionally show the value of incorporating user-selected thresholding. We find that using the optimal threshold for each participant results in an increase in alignment performance in comparison to the fixed threshold. Specifically, we see an increase from 0.650 nMCC to 0.675 nMCC and from 0.668 accuracy to 0.700 accuracy for the IMU \& Throat Mic. algorithm. In addition to these quantitative results, we find the user-selected thresholding to improve performance of the system during pilot testing. In Section~\ref{sec:social_dining_appropriateness}, we additionally discuss how our Phase 3 results demonstrate the usefulness of the assertiveness thresholds for users.

\vspace{-0.15cm}
\subsection{Bite Timing Appropriateness}
\vspace{-0.1cm}
We present results on bite timing appropriateness from our Phase 2 Study in Table~\ref{tab:yesno} for the question ``Would you consider the timing of the bites appropriate?''. In Fig.~\ref{fig:bite_timing}, we present participants' responses to the question ``How was the bite timing of this method on average?''. 

\begin{table}[h]
\vspace{-0.3cm}
\centering
\caption{Phase 2 Responses on Bite Timing Appropriateness}
\vspace{-0.4cm}
\begin{tabular}{|l|c|c|c|}
\hline
\textbf{Context} & \textbf{Fixed Interval} & \textbf{Mouth Open} & \textbf{Wearables} \\
\hline
Individual & 8 (53\%) & 15 (100\%) & 15 (100\%) \\
Social     & 13 (87\%) & 11 (73\%)  & 13 (87\%)  \\
\hline
Overall & 21 (70\%) & 26 (87\%) & \textbf{28 (93\%)} \\
\hline
\end{tabular}
\label{tab:yesno}
\vspace{-0.35cm}
\end{table}

\begin{figure*}[t!]
  \centering
  \vspace{-0.3cm}
  \includegraphics[width = 0.9\textwidth]{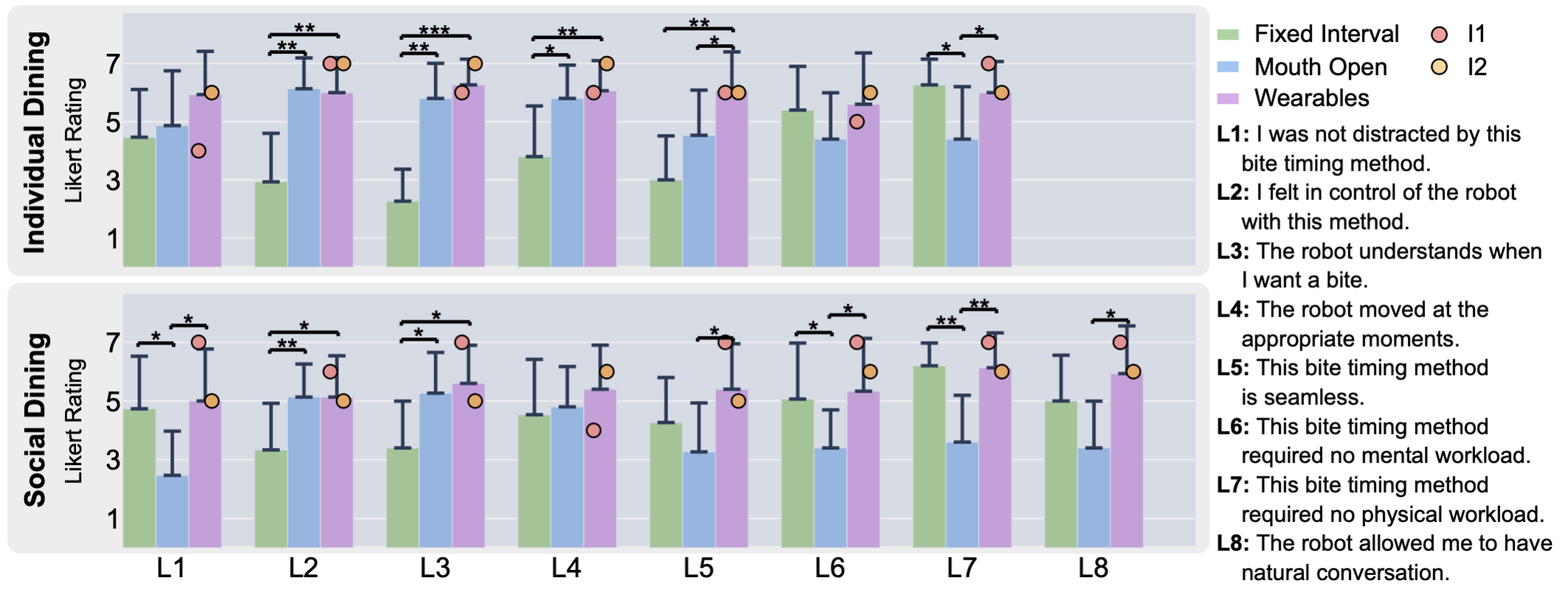}
  \vspace{-0.4cm}
  \caption{Phase 2 and Phase 3 Individual and Social Dining Mean Likert Ratings with error bars representing the standard deviation. Higher scores indicate more favorable outcomes for all items. Asterisks denote statistically significant pairwise differences based on Wilcoxon signed-rank tests with Bonferroni correction ($p < 0.05$ = *, $p < 0.01$ = **, $p < 0.001$ = ***). Prior to pairwise comparisons, a non-parametric Friedman test was used to confirm overall differences among the three methods for each Likert item. Data is additionally included in Table~\ref{tab:individual_FI}-~\ref{tab:social_W}.}
  \Description{TODO}
  \label{fig:combined_likert}
  \vspace{-0.5cm}
\end{figure*}

\vspace{-0.13cm}
\subsubsection{Individual Dining} 
As seen in Table~\ref{tab:yesno}, for individual dining, we find that the Wearables method performs significantly better than both baseline methods. All 15 Phase 2 participants rated the timing of the Wearables method as appropriate, comparable to the Mouth Open method, while only 8/15 participants (53\%) rated the Fixed Interval method as appropriate. In our Phase 3 study, both participants rated the timing of the Wearables method as appropriate, demonstrating the ability of our algorithm to generalize to both another robotic system and individuals with motor impairments. 

As shown in Fig.~\ref{fig:bite_timing}, for the Phase 2 study, the Wearables method received 13 Right Time responses and only 2 Slightly Early responses. In contrast, the next best method, Mouth Open, received only 9 Right Time responses. The Fixed Interval method received no Right Time responses and was consistently rated as Slightly Late, Late, or Very Late. Meanwhile, in Phase 3 studies, both individuals with motor impairments selected ``Right Time'' for the Wearables method. These results highlight the clear timing advantages of our method in individual settings. By utilizing wearable signals, the system is able to dynamically initiate feeding when chewing subsides and to pause when chewing is detected, enabling bite timing that aligns more closely with the user’s readiness. In contrast, the Fixed Interval method performs worst, feeding every 45 seconds regardless of user behavior. Even if the interval were shortened, it would still lack responsiveness to varying chewing durations across different foods. Interestingly, the Mouth Open method, despite being manually triggered, does not outperform Wearables. This is likely due both to the inherent delay between a participant opening their mouth and the robot completing its trajectory to deliver the bite, as well as the need for participants to provide repetitive input for each bite.

Qualitative feedback further supports the quantitative findings. Several participants highlighted that the Wearables method aligned naturally with their preferred eating pace. For instance, P2 said ``In terms of the individual portion, I liked this [Wearables] timing the best because it didn't force me to wait for another bite like the other ones, it kind of naturally matches my pace, the other ones felt a little slow or fast at times.'' The two participants with motor impairments also strongly endorsed this approach. I2 remarked, ``Yes, I like it...I didn't have to think about it...The timing was great, it was...I loved it really.''

\vspace{-0.15cm}
\subsubsection{Social Dining} 
\label{sec:social_dining_appropriateness}
As seen in Table~\ref{tab:yesno}, from our Phase 2 data for social dining, we find that the Wearables and Fixed Interval methods perform similarly in terms of appropriateness with 13/15 (87\%) of participants finding the methods appropriate. Mouth Open performed the worst for social dining with only 11/15 (73\%) participants finding the method appropriate. In our Phase 3 study, for social dining, both participants once again rated the timing of the Wearables method as appropriate. 

As shown in Fig.~\ref{fig:bite_timing}, responses exhibit greater variability across all methods compared to individual dining, reflecting the increased variability in social dining settings due to conversation. Both the Wearables and Fixed Interval methods received 8 \textit{Right Time} responses. However, the remaining Wearables responses were more tightly clustered around \textit{Right Time}, with only 2 rated as early and 1 as late, whereas Fixed Interval received 1 early and 3 late ratings. Meanwhile, in our Phase 3 studies, I1 selected ``Right Time'' and I2 selected ``Slightly Early'' for the Wearables method. 

Qualitative responses also reflect mixed preferences between the Fixed Interval and Wearables methods. Some participants found Fixed Interval disruptive, noting it brought the robot ``at random points during the conversation'' (P15), while others valued it as ``easier to anticipate'' (P3). For the Wearables method, some participants appreciated its adaptive nature, such as P2, who remarked, ``it would wait for me if I started chewing or talking after I signaled that I was ready to eat. I didn’t have to control the conversation for a pause around the time that the bite would be ready.'' In contrast, P8 felt the robot was ``smothering me with food when I was hoping to talk'' (P8). These differing perspectives highlight the value of incorporating the user-controlled assertiveness threshold. In our Phase 3 study, where the 2 participants were allowed to select their own thresholds, they chose values at opposite ends of the spectrum, and both reported positive and enjoyable experiences. I1 said he liked the method during social dining ``because it just let me have a conversation and eat at the same time'' while I2 said it was ``almost seamless.''

\vspace{-0.25cm}
\subsection{Likert Items}
\vspace{-0.1cm}
For the Likert items described in Section~\ref{sec:phase2_methods}, from the Phase 2 studies, we present the mean and standard deviation for each item for individual and social dining in Fig.~\ref{fig:combined_likert}.  We additionally present results from statistical testing in the plot. We specifically use the non-parametric Friedman test to assess overall differences among the three methods for each item. Significant differences ($p < .05$) are found for all items except \textit{L4: Appropriate Movement} in the social dining context. For items showing significant overall effects, we conduct post-hoc Wilcoxon signed-rank tests for all pairwise method comparisons, applying the conservative Bonferroni correction factor of 3. In the plot, we also include the data points for the Wearables method from the Phase 3 study with I1 and I2.

\vspace{-0.15cm}
\subsubsection{Individual Dining} 
As seen in Fig.~\ref{fig:combined_likert}, the Wearables method received strong Likert ratings by both Phase 2 and Phase 3 participants in the individual dining condition across all items, with statistically higher ratings for \textit{L2: Feeling of Control}, \textit{L3: Robot Understanding}, \textit{L4: Appropriate Movement}, \textit{L5: Seamlessness}, and \textit{L7: No Physical Workload} compared to at least one baseline. 

Wearables achieved similar \textit{Feeling of Control} as Mouth Open but with statistically lower \textit{Physical Workload}. Both participants with impairments gave a score of 7 (Strongly Agree) for \textit{L2: Feeling of Control} and score of 7 (Strongly Agree) and 6 (Agree) for \textit{L7: No Physical Workload}. This is notable, as prior studies report a trade-off where autonomy often reduces user feeling of control~\cite{javdani2018,bhattacharjee2020}. In contrast, our method balanced autonomy and user intent, reducing workload while preserving control. P6 described the system as one that ``understood my natural body movements and I did not have to exert additional effort.'' P15 similarly emphasized efficiency, stating it ``required the least work on my part and had the best timing. I didn’t have to open my mouth or wait unnecessarily long.'' The participants with impairments also felt similarly with I1 saying, ``I do like it. Cause I’m in control of the meal, when I was gonna take the bite, when I wasn’t.'' 

Wearables was statistically similar to Mouth Open on \textit{L3: Robot Understanding} and \textit{L4: Appropriate Movement}, showing that participants perceived it as equally responsive despite relying on subtle cues like chewing and motion rather than explicit mouth opening. This performance underscores its ability to interpret nuanced behavior without any added workload. Lastly, Wearables was rated more seamless than both baselines on \textit{L5: Seamlessness} with statistical significance, indicating a smoother, more intuitive experience. This supports the idea that leveraging natural behaviors reduces friction in human–robot interaction, enhancing comfort without sacrificing control.

\vspace{-0.25cm}
\subsubsection{Social Dining} 
As seen in Fig.~\ref{fig:combined_likert}, the Wearables method again received strong Likert ratings for all items for both individuals with and without impairments. The Wearables method received the highest ratings for \textit{L1: Not Distracted} and \textit{L8: Natural Conversation}, statistically outperforming Mouth Open. This suggests the method lets the robot feed without disrupting conversation, unlike explicit cues such as mouth opening. P3 noted, ``the actions for pausing aligned with social cues where I'd rather not be disrupted.'' In contrast, Fixed Interval showed no statistically significant difference on \textit{L8: Natural Conversation} compared to Mouth Open. We find similar, strong responses for the Wearables method from the 2 individuals with impairments, with I1 provided a response of 7 (Strongly Agree) to \textit{L8: Natural Conversation} while I2 responded with a 6 (Agree). I1 remarked, ``It didn’t seem to take away from the conversation, so it was just kind of part of that environment I guess?...It just seemed to not interrupt the conversation.'' 

For Phase 2 participants, Wearables also matched Mouth Open on \textit{L2: Feeling of Control} and \textit{L3: Robot Understanding}, while statistically outperforming Fixed Interval. This indicates that even in dynamic social contexts, the wearable sensing approach remains responsive and aligned to user intent. Finally, Wearables was rated as more seamless on \textit{L5: Seamlessness} and yielded statistically lower \textit{L6: Mental Workload} and \textit{L7: Physical Workload} than Mouth Open. P1 captured this sentiment succinctly: ``It...recreated the way I usually eat.'' Notably, Wearables matched Fixed Interval, a fully autonomous method, on workload while offering more adaptive behavior.

\begin{figure*}[t!]
  \vspace{-0.3cm}
  \centering
  \includegraphics[width = 0.9\textwidth]{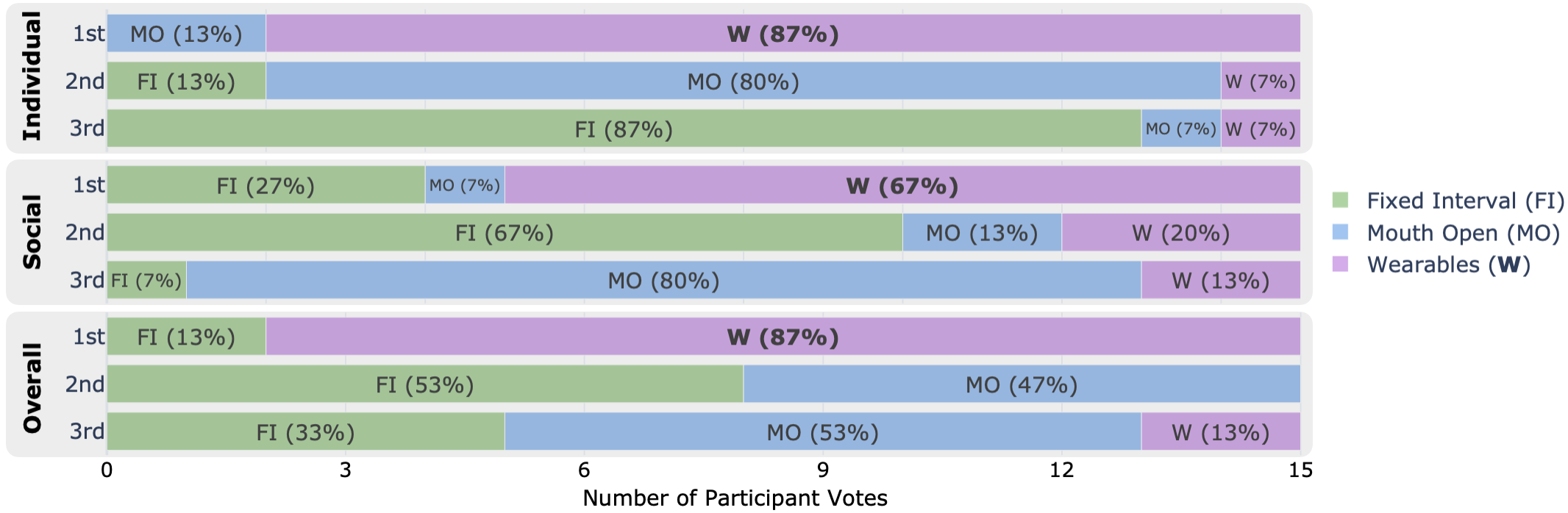}
    \vspace{-0.35cm}
  \caption{Phase 2 preference rankings for the three methods. The Wearables method received the most 1st-place rankings across individual, social, and overall contexts.} 
  \Description{TODO}
  \label{fig:preference_rankings}
  \vspace{-0.55cm}
\end{figure*}

\vspace{-0.3cm}
\subsection{Ranking and Caregiver Similarity}
\vspace{-0.1cm}
As detailed in Section~\ref{sec:phase2_methods}, Phase 2 participants blindly ranked the methods after individual and social dining. As seen in Fig.~\ref{fig:preference_rankings}, the Wearables method was strongly preferred, receiving 87\% of first-place votes in the individual setting and 67\% for social dining. In unblinded overall rankings, it received the majority of first-place votes (87\%). Fixed Interval was generally the next most preferred, while Mouth Open was least favored, particularly in the social setting where participants noted it disrupted conversational flow. We likely observe less first place votes for WAFFLE in the social setting in comparison to the individual setting as the limited number of trials and conversational variability made comparative evaluation more challenging. Nevertheless, our results show that WAFFLE provided users with statistically higher feeling of control, more robot understanding, and more adaptability than the Fixed Interval method in the social setting. 

When asked which method most closely resembled how a caregiver would decide when to feed, a majority (9/15) of Phase 2 participants selected the Wearables method. In comparison, 3 chose Mouth Open, 1 chose Fixed Interval, and 2 participants mentioned both Wearables and Mouth Open. P5 summarized why the Wearables method was the closest to a caregiver saying, ``it wouldn't give the bite when I clearly didn't want it, and it would adjust to the environment to decide when to give the next bite, like a person would.'' We highlight the cues participants find most important during feeding in Appendix~\ref{appendix:feeding_cues}. These Phase 2 results were strongly reinforced by the participants with motor impairments in the Phase 3 study. When asked if the Wearables method is similar to a caregiver, I1 said, ``I would say yes. Because when I wanted a bite, I got it.'' I2 expressed even stronger support, saying ``way better, way better than they would do it...way better just to be truthful.'' She elaborated that she often has to instruct caregivers directly, saying, ``watch for me chewing... if I’m still chewing, you don’t need to pick the fork up,'' and noted that some of them don't pay attention. She elaborated saying ``It drives me crazy when I'm talking to somebody...and they go to give me a bite.'' Lastly, she added that caregivers can be easily distracted, saying``If we’re watching TV, forget about it. This is so much better, cause if we’re watching TV, they don’t watch me as much as they watch the TV...So yeah this is great.'' These results suggest that the Wearables method aligned with caregiver-like behavior and at times even exceeded caregiver performance.

\vspace{-0.2cm}
\section{Adoption and Future Work}
\vspace{-0.15cm}
In this work, we presented WAFFLE, a wearable-based system for estimating bite timing in robot-assisted feeding. By leveraging natural signals from wearables, WAFFLE enables intuitive and socially appropriate interactions, responding in real time to passive user cues such as chewing, speaking, and head movements. Across studies with participants with and without motor impairments, WAFFLE consistently outperformed baselines in feeling of control, robot understanding, workload, and preference, supporting our hypothesis that robots can achieve unobtrusive and generalizable bite timing by reacting to natural cues. The method also generalized effectively across many participants, robot platforms, trajectories, foods, and dining scenarios, underscoring its robustness and adaptability. In our Phase 3 study, we found this generalization extended to unseen, real-world contexts; as examples, I1 successfully used the system while watching TV at home during individual dining, and the robot appropriately paused when I2 sneezed during the study.

These results indicate strong likelihood of adoption of our interface by users in future commercial robotic feeding systems. This is further reinforced by the positive reception from participants with motor impairments, who had prior experience using feeding robots with web interfaces, and reported that our system was similar to or better than their human caregivers. To further assess potential adoption, we qualitatively evaluated the comfort of the two wearable devices, as wearability plays a crucial role in sustained use. We found that the vast majority of participants considered both the IMU-equipped glasses and the throat microphone comfortable to wear during dining, with additional details provided in Appendix~\ref{appendix:wearability}.

Given this promising outlook, it is worth noting that wearable sensing with IMUs and throat microphones is increasingly practical for real-world deployment. IMUs are already integrated into commercially available smartglasses and throat microphones currently have utility as general-purpose microphones. In terms of robot-assisted feeding specifically, these wearables could also offer multiple use cases beyond bite timing, such as tracking food intake or sensing spasms or choking. The wearables are privacy preserving and the throat microphone does not pick up ambient speech or sounds, allowing for reliable operation in noisy environments. The wearables are unobtrusive, further supporting adoption in daily life. These insights are in line with past work that has shown that wearable interfaces offer key benefits over conventional interfaces for caregiving robots, enabling intuitive, accessible, and unobtrusive control~\cite{padmanabha2024independence, padmanabha2023hat, padmanabha2025towards, yang2023high}. We find comparable advantages in our study, with the proposed Wearables method outperforming the baselines that are more disruptive because they rely on screens, assistive input devices, or require the robot to approach from the front or the user to face the robot each time they want a bite. In contrast, WAFFLE reduces disruption and supports natural interaction in both individual and social dining situations. 

Future work should include more testing with individuals with impairments, particularly through longitudinal studies in diverse dining environments to reveal long-term trends in usage. Another promising direction is to incorporate additional cues identified by participants, such as eye gaze.

\bibliographystyle{ACM-Reference-Format}
\balance
\bibliography{bibliography}

\clearpage

\appendix
\setcounter{table}{0}
\renewcommand{\thetable}{A\arabic{table}}
\renewcommand{\thefigure}{A\arabic{figure}}
\setcounter{figure}{0}

\section{Appendices}

\subsection{Devices}
\label{appendix:devices}
For the IMUs, we use open-sourced, custom-built printed circuit boards~\cite{SMLPosey2025} consisting of the CEVA Technologies BNO086 IMU. The IMU boards are wireless and communicate over Bluetooth with a hub (Nordic Semiconductor NRF5340-DK) which forwards IMU readings over wired serial communication to the laptop. For the throat microphone, we use the Lsgoodcare 3.5MM Plug Adjustable Throat Microphone. 

\subsection{Modeling Additional Details}
\label{appendix:modeling}

We use a regression model to estimate the time until the next bite rather than directly predicting binary proceed/stop commands for two key reasons. First, this approach yields a more generalizable model that can support various deployment strategies, such as converting predictions into binary commands or using them directly to control the number of seconds the robot should pause until progressing through the feeding trajectory. Second, participant-issued progress/stop commands were often noisy and inconsistent, even within the same individual, making it difficult for the model to learn associations between wearable signals and robot motion. For example, a participant might advance the robot mid-chew in one trial but wait until finishing chewing in another which are both valid behaviors for appropriate bite timing.

Our data and model pipeline is explained with additional detail as follows. First, 1s windows of 3-axis IMU accelerations and throat microphone data are resampled using 1-D interpolation to 200 Hz and 100 Hz respectively. We specifically select short 1s windows as this minimizes latency and allows our feeding robot to be very reactive to human behavior and cues. Next, we split each 1s sample into two 500 ms low-level windows. For each low-level window, for each of the 3 acceleration axes of the IMU and the throat microphone signal, we extract 6 hand-picked features from the acceleration: maximum, minimum, mean, standard deviation, range, and root mean square (RMS). We concatenate the features from both sensing modalities and normalize them to zero mean and unit variance. In total, there are 48 features (2 low level windows $\times$ 4 axes $\times$ 6 features per axis). The concatenated, normalized features, $f \in \mathbb{R}^{48}$ are inputted to our bite timing ML model, which is a multi-layer perceptron with three fully connected layers of sizes (128, 64, and 1). ReLU activations and dropout with a probability of 0.1 are applied after the first and second layers. The final layer directly outputs the predicted time (s), $\hat{y}$, until the next bite. We train the model using ground truth labels of time until the next bite, collected from our human study. We observe that human behavior more than 10s before a bite, such as chewing, talking, or head movements, is similar to behavior around the 10 second mark. To limit the influence of these distant and less informative events during model training, we cap all ground truth labels at 10s. This reduces influence on the loss from larger labels that will not improve bite timing prediction. The model is trained using mean absolute error (MAE) loss and the Adam optimizer for 100 epochs with a learning rate of 0.0001 and batch size of 128. Our bite timing model generates predictions at 2 Hz. The predictions are thresholded using the user selected assertiveness threshold and the binary progress/stop command is sent to the robot at 2 Hz.

\subsection{Study Setup Additional Details}
\label{appendix:study_setup}

\begin{figure*}[tbp]
  \centering
  \includegraphics[width = \textwidth]{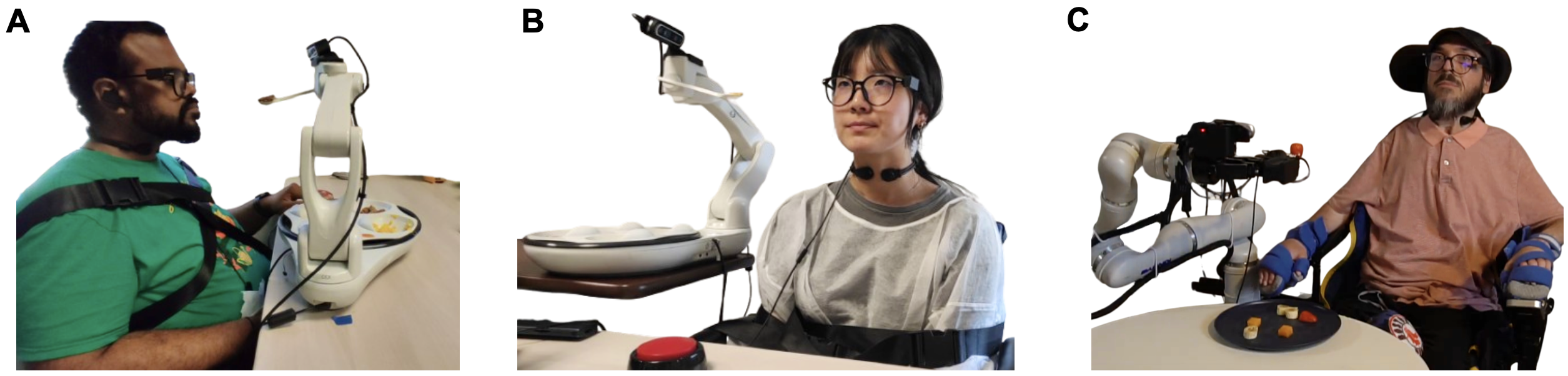}
  \caption{Study Setup for the Three Phases. \textbf{A.} Phase 1 Study Setup. In this study, we collect bite timing data from 14 participants with no motor impairments with the Obi robot feeding from the front of the participant. \textbf{B.} Phase 2 Study Setup. In this study, we compare our wearables approach, WAFFLE, to two baselines with 15 participants without motor impairments with the Obi robot feeding from the side of the participant. \textbf{C.} Phase 3 Study Setup. In this study, we evaluate our wearables approach with 2 participants with motor impairments with the FEAST robot~\cite{jenamani2025feast} feeding from the front of the participant.}
  \Description{TODO}
  \label{fig:study_3_phases}
\end{figure*}

\begin{table*}[h]
\centering
\caption{Overview of the Three Human Studies}
\vspace{-1em}
\begin{tabular}{|l|c|c|c|}
\hline
\textbf{} & \textbf{Phase 1} & \textbf{Phase 2} & \textbf{Phase 3} \\
\hline
\textbf{Type of Study} & Data Collection & Evaluation & Evaluation \\
\hline
\textbf{Number of Participants}  & 14 & 15 & 2 \\
\hline
\textbf{Demographic}  & No Motor Impairments & No Motor Impairments & Motor Impairments  \\
\hline
\textbf{Robot Type} & Obi~\cite{Obi} & Obi~\cite{Obi}  & FEAST~\cite{jenamani2025feast}\\
\hline
\textbf{Robot Positioning} & Front & Side  & Side\\
\hline
\textbf{Feeding Trajectory} & Front & Side  & Front  \\
\hline
\textbf{Foods} & Assorted & Assorted & Assorted Fruits \\
\hline
\textbf{Settings} & Individual \& Social &  Individual \& Social & Individual \& Social \\
\hline
\textbf{Bite Timing Methods} & Participant Controlled \& WoZ & Wearables \& 2 Baselines & Wearables \\
\hline
\end{tabular}
\label{tab:study_comparison}
\end{table*}

We visually show the study setup for the three phases in Fig.~\ref{fig:study_3_phases} and directly compare them in Table~\ref{tab:study_comparison}.

The following questions and Likert items were asked to participants after testing each method in the Phase 2 study:
\begin{itemize}
    \item Would you consider the timing of the bites appropriate? (Yes/No)
    \item How was the bite timing of this method on average? (Very Early, Early, Slightly Early, Right Time, Slightly Late, Late, Very Late)
    \item L1: I was not distracted by this bite timing method.
    \item L2: I felt in control of the robot with this method.
    \item L3: The robot understands when I want a bite.
    \item L4: The robot moved at the appropriate moments.
    \item L5: This bite timing method is seamless.
    \item L6: This bite timing method required no mental workload.
    \item L7: This bite timing method required no physical workload.
    \item L8: For social dining only: The robot allowed me to have natural conversation. 
\end{itemize}

At the end of the Phase 2 study, the participants answer the following qualitative questions: 
\begin{itemize}
    \item Why did you prefer your highest ranked method? Please elaborate and compare the methods for both the individual and social settings.
    \item Which methods did you feel in control while using? Why or why not?
    \item Which method had the most appropriate bite timing? Elaborate on why it had the most appropriate timing and compare it to the other methods.
    \item What cues do you think a caregiver or you use to know when to feed a care recipient or yourself a bite?
    \item Which method was closest to how a caregiver or you would perform this task? Please elaborate on why.
    \item Were the glasses comfortable to wear while eating? If not, please explain.
    \item Was the throat microphone comfortable to wear while eating? If not, please explain.
\end{itemize}

\subsection{Social Dining Conversation Topics}
\label{appendix:socialdining}

Similar to Ondras et al.~\cite{ondras2022human}, we utilize the following conversation starters for the social dining portion of the study: 
\begin{itemize}
    \item Who is your favorite singer and why?
    \item What are you studying?
    \item What is your favorite food and why?
    \item What is your favorite color and why?
    \item Do you give back or volunteer with any organizations?
    \item What are your favorite writers and books?
    \item Do you have any pets and if so, what are they?
    \item What sports do you play or watch and why?
    \item What is your favorite movie and why?
    \item Who is your favorite actor and why?
    \item Which languages do you speak and which ones do you want to learn?
    \item What was your favorite vacation?
    \item What are your hobbies?
\end{itemize}

To foster a more natural conversation, we encourage participants to ask the researcher questions in return. Additionally, we have the researcher eat food to simulate a shared meal experience.

\subsection{Wizard of Oz Guidelines}
\label{appendix:wizardofoz}
To mitigate participant suspicion, the researcher uses a video feed from the RealSense camera displayed on the companion robot to get a clear view of the participant's face during data collection. These guidelines are listed in order of priority as sometimes cues can conflict; for example, the person may still be chewing while also glancing at the robot to indicate they want another bite.  

\subsubsection{Individual}
Listed in order of priority, so in the case of a conflict, higher ones should be adhered to before lower ones:

\begin{enumerate}
    \item Move forward: Participant starts glancing toward the robot or the camera
    \item Stop: Participant’s body language or other forms of communication indicate they do not want the bite (e.g., leaning backwards, turning their head away from the spoon, etc.)
    \item Move forward: Participant is not chewing
    \item Move forward: Participant is giving indication that they are almost done chewing, such as:
    \begin{itemize}
        \item Visibly or audibly chewing significantly more lightly than previously
        \item Licking their mouth area
        \item Visibly or audibly swallowing
    \end{itemize}
    \item Stop: Participant is chewing
\end{enumerate}

\subsubsection{Social}
Listed in order of priority, so in the case of a conflict, higher ones should be adhered to before lower ones:

\begin{enumerate}
    \item Stop: Participant’s body language or other forms of communication indicate they do not want the bite (e.g., leaning backwards, turning their head away from the spoon, etc.)
    \item Move forward: Participant is giving indication that they are almost done talking, such as:
    \begin{itemize}
        \item Posing a question to the dining partner
        \item Trailing off
    \end{itemize}
    \item Stop: Participant is talking
    \item Stop: Dining partner is giving indication that they are almost done talking, such as:
    \begin{itemize}
        \item Posing a question to the participant
        \item Trailing off
    \end{itemize}
    \item Move forward: Dining partner is talking
    \item Move forward: Participant is not chewing
    \item Move forward: Participant is giving indication that they are almost done chewing, such as:
    \begin{itemize}
        \item Visibly or audibly chewing significantly more lightly than previously
        \item Licking their mouth area
        \item Visibly or audibly swallowing
    \end{itemize}
    \item Stop: Participant is chewing
\end{enumerate}

\subsection{Food Options}
\label{appendix:foods}

For Phase 1 studies, we allow participants to pick a soft, medium, and hard food and an additional wildcard food from any of the following 3 categories to fill the 4 bowls of the Obi robot:
\begin{itemize}
    \item Soft: yogurt, applesauce, tomato soup 
    \item Medium: rice, mac and cheese, corn, mashed potatoes
    \item Hard: pretzels, cheerios, granola, goldfish
\end{itemize}

For Phase 2 studies, we all participants to pick 2 medium and 2 hard foods. We exclude soft foods from this study as there is increased risk of food spillage due to feeding from the side. The foods chosen by participants in Phase 1 and Phase 2 are provided in Table~\ref{tab:phase1_foods} and Table~\ref{tab:phase2_foods}. For both Phase 1 and Phase 2 studies, a small display monitor is placed on the table in front of the user to indicate which bowl the robot will be scooping from next. The fruits chosen by participants in Phase 3 are provided in the main text.

\begin{table*}[ht!]
\centering
\caption{Phase 1 Food Selections}
\vspace{-0.4cm}
\label{tab:phase1_foods}
\begin{tabular}{lcccc}
\toprule
\textbf{Participant} & \textbf{Hard Food} & \textbf{Medium Food} & \textbf{Soft Food} & \textbf{Wildcard} \\
\midrule
1 & pretzels & mac and cheese & applesauce & granola \\
2 & pretzels & rice & yogurt & cheerios \\
3 & pretzels & rice & applesauce & granola \\
4 & granola & mashed potatoes & yogurt & goldfish \\
5 & goldfish & rice & applesauce & mac and cheese \\
6 & cheerios & rice & yogurt & mac and cheese \\
7 & granola & mashed potatoes & yogurt & pretzels \\
8 & granola & mac and cheese & tomato soup & mashed potatoes \\
9 & cheerios & corn & applesauce & granola \\
10 & goldfish & corn & yogurt & tomato soup \\
11 & granola & mashed potatoes & yogurt & applesauce \\
12 & cheerios & mashed potatoes & yogurt & pretzels \\
13 & goldfish & corn & yogurt & pretzels \\
14 & pretzels & rice & applesauce & granola \\
\bottomrule
\end{tabular}
\end{table*}

\begin{table*}[ht!]
\centering
\caption{Phase 2 Food Selections}
\vspace{-0.4cm}
\label{tab:phase2_foods}
\begin{tabular}{lcccc}
\toprule
\textbf{Participant} & \textbf{Hard Food} & \textbf{Medium Food} & \textbf{Hard Food} & \textbf{Medium Food} \\
\midrule
1 & cheerios & rice & granola & mac and cheese \\
2 & granola & mashed potatoes & cheerios & mac and cheese \\
3 & cheerios & mac and cheese & granola & rice \\
4 & pretzels & mac and cheese & cheerios & mashed potatoes \\
5 & pretzels & mashed potatoes & granola & mac and cheese \\
6 & cheerios & mashed potatoes & goldfish & mac and cheese \\
7 & pretzels & rice & granola & mashed potatoes \\
8 & goldfish & mashed potatoes & cheerios & mac and cheese \\
9 & goldfish & rice & pretzels & mac and cheese \\
10 & cheerios & mashed potatoes &  pretzels& rice \\
11 & pretzels & rice & granola  & mac and cheese \\
12 & granola & rice & goldfish & mashed potatoes \\
13 & goldfish & mashed potatoes & cheerios & mac and cheese \\
14 & goldfish & mashed potatoes & cheerios & mac and cheese \\
15 & goldfish & rice & pretzels & mashed potatoes \\
\bottomrule
\end{tabular}
\end{table*}

\subsection{Participant Demographics}

\subsubsection{Phase 1}
\label{appendix:phase1_demographics}
\begin{figure*}[tbp]
  \centering
  \includegraphics[width = \textwidth]{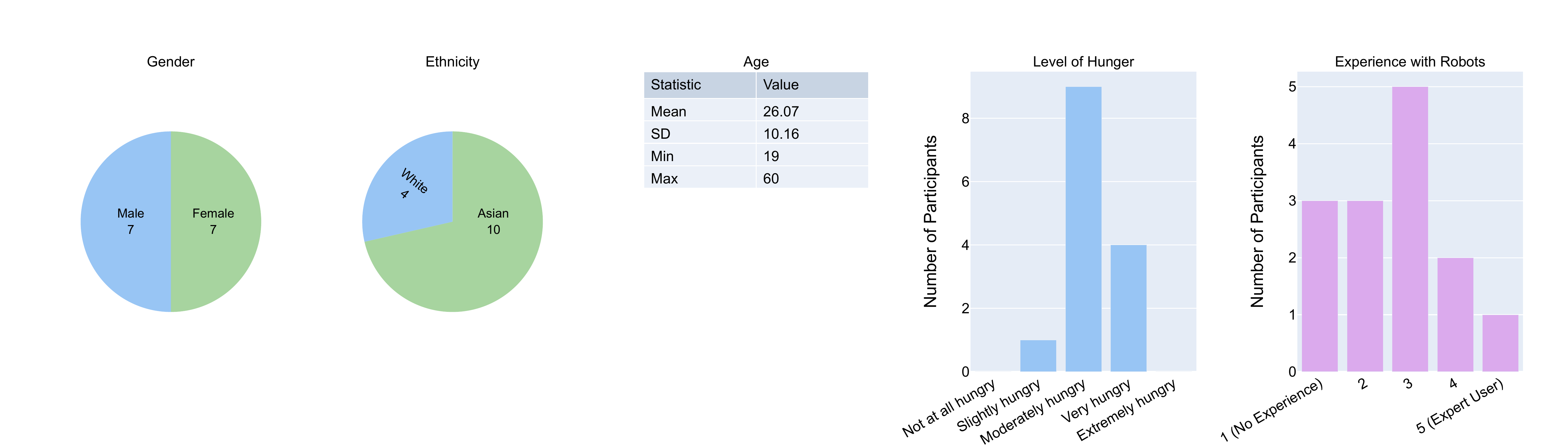}
  \caption{Phase 1 (n=14) Demographics}
  \Description{TODO}
  \label{fig:phase1_dem}
\end{figure*}

In addition to the demographics detailed in the main text, participants were asked how much experience they have with controlling a robot and their level of hunger. Three additional participants were excluded, two because they took very few bites as they were not hungry during the session, and one due to sensor connection issues. We visually show participant demographics from our Phase 1 studies in Fig~\ref{fig:phase1_dem}. 

\subsubsection{Phase 2}
\label{appendix:phase2_demographics}
\begin{figure*}[tbp]
  \centering
  \includegraphics[width = \textwidth]{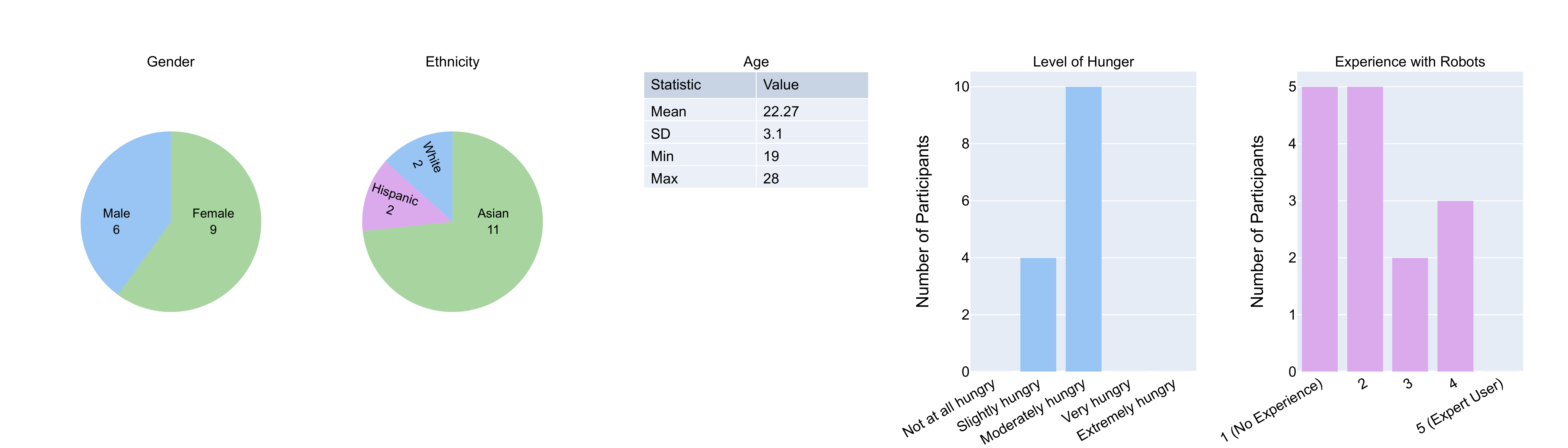}
  \caption{Phase 2 (n=15) Demographics}
  \Description{TODO}
  \label{fig:phase2_dem}
\end{figure*}

Similar to the Phase 1 study, participants were also asked how much experience they have with controlling a robot and their level of hunger. We visually show participant demographics from our Phase 2 studies in Fig~\ref{fig:phase2_dem}.

\subsubsection{Phase 3}
\label{appendix:phase3_demographics}
The majority of Phase 3 demographics information is presented in the main text in Section~\ref{sec:phase3_methods}. Additionally, both participants with motor impairments indicated their level of hunger was ``Moderately Hungry''.

\subsection{Metrics using Different Fixed Assertiveness Thresholds}
\label{appendix:fixed_thresholds}

We include the accuracy and nMCC metrics for fixed assertiveness threshold values from 4 seconds to 8 seconds in Table~\ref{tab:fixed_thresholds}. We find that the threshold of 6 seconds outperforms all other threshold values. 

\begin{table*}[tbp]
\centering
\caption{Accuracy and nMCC for Fixed Assertiveness Threshold Values from 4s to 8s}
\vspace{-0.4cm}
\begin{tabular}{lccccc}
\toprule
\textbf{Threshold Value} & \textbf{4s} & \textbf{5s} & \textbf{6s} & \textbf{7s} & \textbf{8s} \\
\midrule
Accuracy & $0.608 \pm 0.146$ & $0.627 \pm 0.115$ & $\mathbf{0.668 \pm 0.066}$ & $0.552 \pm 0.098$ & $0.471 \pm 0.144$ \\
nMCC     & $0.566 \pm 0.056$ & $0.614 \pm 0.046$ & $\mathbf{0.650 \pm 0.082}$ & $0.589 \pm 0.053$ & $0.554 \pm 0.064$ \\
\bottomrule
\end{tabular}
\label{tab:fixed_thresholds}
\end{table*}

\subsection{Wearability}
\label{appendix:wearability}
Participant responses on wearability of the two devices from the Phase 2 and Phase 3 studies are provided in Appendix~\ref{appendix:qualitative_responses} and \ref{appendix:phase3_questions_responses}. We find that the vast majority of participants considered both the IMU-equipped glasses and the throat microphone comfortable to wear. 6/15 participants initially found the throat microphone uncomfortable or distracting to wear, but all clarified, without prompting, that they stopped noticing it once the study began. One participant, however, mentioned that they would not want to wear it for an extended period. For users who may find the throat microphone undesirable, we believe IMU-only models could provide a reasonable alternative based on our bite timing results in Section~\ref{sec:model_performance}, though further testing is needed to confirm this. 

From the Phase 3 studies, we find that both participants with motor impairments reported that the IMU-equipped glasses and throat microphone were generally comfortable to wear across both individual and social dining scenarios. Both noted that any initial awareness of the devices diminished over time, with I2 mentioning that while the throat microphone initially felt ``awkward'' or ``a little snug,'' it was not uncomfortable. Similarly, both participants described the IMU on the glasses as unobtrusive, with I2 commenting that they ``forgot it was there'' after the start of the session. These findings align closely with our Phase 2 results, suggesting that both sensing modalities are well-accepted even in real-world, in-home settings.

\subsection{Feeding Cues}
\label{appendix:feeding_cues}
At the end of the Phase 2 and 3 studies, we additionally gauged participants on what cues a caregiver or they themselves would use to decide when to deliver a bite of food, and which bite timing method best reflected that behavior. When describing the cues they would rely on, 8 of the Phase 2 participants mentioned chewing or stopping chewing and 7 participants mentioned explicit gestures such as opening the mouth, head turning, or nodding. Other cues included talking or stopping talking (6 participants), swallowing (3 participants), looking at the food or caregiver (2 participants), and being still (1 participant). 

From the Phase 3 studies, both participants with motor impairments similarly emphasized chewing status as an important cue for appropriate bite timing. I1 explained that a caregiver would likely rely on ``when I was done chewing'' or ``if I turn toward them.'' In social dining, I1 added that pauses in activity could also serve as cues: ``if I stop talking, maybe they’re like `oh, he wants a bite'.''

I2 reported giving similar guidance to caregivers: ``watch for me chewing... if I’m still chewing, you don’t need to pick the fork up.'' They also described frustration when bites were offered too soon and stressed that feeding should not interrupt conversation: ``It drives me crazy when I’m talking... and they go to give me a bite.'' Finally, they noted that caregivers can sometimes miss cues: ``If we’re watching TV... they don’t watch me as much as they watch the TV... so this is better.''

\subsection{Statistical Testing}
\label{appendix:statistical_testing}
\subsubsection{Friedman Test}
P-values and Friedman statistics are provided for individual and social dining in Table~\ref{table:friedman_individual} and Table~\ref{table:friedman_social} respectively. 

\begin{table}[H]
\centering
\caption{Friedman Test Results for Individual Dining}
\begin{tabular}{lcc}
\toprule
Likert Item &  Friedman Statistic &  p-value \\
\midrule
         L1 &            7.541667 & 0.023033 \\
         L2 &           21.166667 & 0.000025 \\
         L3 &           24.111111 & 0.000006 \\
         L4 &           18.285714 & 0.000107 \\
         L5 &           16.472727 & 0.000265 \\
         L6 &            7.409091 & 0.024611 \\
         L7 &            8.711111 & 0.012835 \\
\bottomrule
\end{tabular}
\label{table:friedman_individual}
\end{table}

\begin{table}[H]
\centering
\caption{Friedman Test Results for Social Dining}
\begin{tabular}{lcc}
\toprule
Likert Item &  Friedman Statistic &  p-value \\
\midrule
         L1 &           14.509091 & 0.000707 \\
         L2 &           13.377778 & 0.001245 \\
         L3 &           10.619048 & 0.004944 \\
         L4 &            2.039216 & 0.360736 \\
         L5 &           11.230769 & 0.003641 \\
         L6 &            9.529412 & 0.008525 \\
         L7 &           21.115385 & 0.000026 \\
         L8 &           14.333333 & 0.000772 \\
\bottomrule
\end{tabular}
\label{table:friedman_social}
\end{table}

\subsubsection{Wilcoxon Signed-Rank Test}
Post hoc results using the Wilcoxon signed-rank tests  are provided for individual and social dining in Table~\ref{table:wilcoxon_individual} and Table~\ref{table:wilcoxon_social} respectively. 

\begin{table*}[tbp]
\centering
\caption{Wilcoxon Signed-Rank Test Results for Individual Dining}
\vspace{-0.4cm}
\begin{tabular}{lrrr}
\toprule
Likert Item &  Fixed Interval - Mouth Open &  Fixed Interval - Wearables &  Mouth Open - Wearables \\
\midrule
         L1 &                     1.666192 &                    0.142750 &                0.540442 \\
         L2 &                     0.004709 &                    0.002705 &                2.017295 \\
         L3 &                     0.002478 &                    0.000183 &                0.572961 \\
         L4 &                     0.013956 &                    0.004047 &                1.252239 \\
         L5 &                     0.058530 &                    0.004760 &                0.036035 \\
         L6 &                     0.322214 &                    2.393876 &                0.073205 \\
         L7 &                     0.021231 &                    1.853910 &                0.017029 \\
\bottomrule
\end{tabular}
\label{table:wilcoxon_individual}
\end{table*}

\begin{table*}[tbp]
\centering
\caption{Wilcoxon Signed-Rank Test Results for Social Dining}
\vspace{-0.4cm}
\begin{tabular}{lrrr}
\toprule
Likert Item &  Fixed Interval - Mouth Open &  Fixed Interval - Wearables &  Mouth Open - Wearables \\
\midrule
         L1 &                     0.014982 &                    2.777954 &                0.030762 \\
         L2 &                     0.009037 &                    0.018879 &                2.875024 \\
         L3 &                     0.032135 &                    0.014780 &                1.186658 \\
         L5 &                     0.381661 &                    0.128704 &                0.010598 \\
         L6 &                     0.033851 &                    2.261657 &                0.041135 \\
         L7 &                     0.002778 &                    2.870818 &                0.004217 \\
         L8 &                     0.066641 &                    0.265522 &                0.019570 \\
\bottomrule
\end{tabular}
\label{table:wilcoxon_social}
\end{table*}

\subsection{Phase 2 and 3 Raw Data}
\label{appendix:raw_data}
Phase 2 raw data for bite timing is provided in Table~\ref{tab:bite_timing}. Phase 2 and 3 Likert item raw data is provided in Table~\ref{tab:individual_FI}-\ref{tab:social_W}.

\begin{table*}[tbp]
\centering
\caption{Timing Appropriateness Ratings by Method and Dining Context (Number of Participants)}
\vspace{-0.4cm}
\label{tab:bite_timing}
\resizebox{\textwidth}{!}{%
\begin{tabular}{llccccccc}
\toprule
\textbf{Context} & \textbf{Method} & \textbf{Very Early} & \textbf{Early} & \textbf{Slightly Early} & \textbf{Right Time} & \textbf{Slightly Late} & \textbf{Late} & \textbf{Very Late} \\
\midrule
\multirow{3}{*}{Individual} 
& Fixed Interval & 0 & 0 & 0 & 0 & 4 & 9 & 2 \\
& Mouth Open     & 0 & 0 & 2 & 9 & 3 & 1 & 0 \\
& Wearables      & 0 & 0 & 2 & 13 & 0 & 0 & 0 \\
\midrule
\multirow{3}{*}{Social} 
& Fixed Interval & 0 & 1 & 2 & 8 & 1 & 3 & 0 \\
& Mouth Open     & 0 & 3 & 1 & 5 & 5 & 1 & 0 \\
& Wearables      & 0 & 2 & 2 & 8 & 2 & 1 & 0 \\
\bottomrule
\end{tabular}%
}
\end{table*}

\begin{table}[H]
\centering
\caption{Individual Fixed Interval Likert Raw Data}
\vspace{-0.4cm}
\label{tab:individual_FI}
\begin{tabular}{lccccccc}
\toprule
\textbf{Participant} & \textbf{L1} & \textbf{L2} & \textbf{L3} & \textbf{L4} & \textbf{L5} & \textbf{L6} & \textbf{L7} \\
\midrule
P1  &   3 &   3 &   2 &   2 &   2 &   5 &   5 \\
P2  &   5 &   3 &   3 &   5 &   3 &   3 &   5 \\
P3  &   2 &   2 &   2 &   3 &   3 &   6 &   6 \\
P4  &   5 &   4 &   5 &   6 &   4 &   3 &   7 \\
P5  &   7 &   5 &   3 &   5 &   4 &   7 &   7 \\
P6  &   5 &   1 &   1 &   5 &   1 &   7 &   7 \\
P7  &   2 &   1 &   1 &   1 &   1 &   7 &   7 \\
P8  &   6 &   6 &   2 &   6 &   6 &   6 &   6 \\
P9  &   3 &   3 &   1 &   3 &   2 &   3 &   7 \\
P10 &   3 &   1 &   1 &   1 &   1 &   4 &   5 \\
P11 &   6 &   2 &   2 &   3 &   4 &   6 &   7 \\
P12 &   6 &   2 &   2 &   3 &   3 &   6 &   6 \\
P13 &   5 &   2 &   3 &   5 &   2 &   6 &   7 \\
P14 &   3 &   3 &   3 &   3 &   5 &   5 &   5 \\
P15 &   6 &   6 &   3 &   6 &   4 &   7 &   7 \\
\bottomrule
\end{tabular}
\end{table}

\begin{table}[H]
\centering
\caption{Individual Mouth Open Likert Raw Data}
\vspace{-0.4cm}
\label{tab:individual_MO}
\begin{tabular}{lccccccc}
\toprule
\textbf{Participant} & \textbf{L1} & \textbf{L2} & \textbf{L3} & \textbf{L4} & \textbf{L5} & \textbf{L6} & \textbf{L7} \\
\midrule
P1  &   3 &   6 &   5 &   5 &   4 &   5 &   3 \\
P2  &   7 &   7 &   7 &   7 &   6 &   6 &   6 \\
P3  &   6 &   6 &   6 &   6 &   6 &   6 &   6 \\
P4  &   3 &   7 &   6 &   7 &   6 &   5 &   7 \\
P5  &   3 &   3 &   3 &   3 &   2 &   5 &   3 \\
P6  &   7 &   7 &   7 &   7 &   3 &   2 &   2 \\
P7  &   3 &   7 &   6 &   6 &   6 &   6 &   5 \\
P8  &   6 &   6 &   6 &   6 &   5 &   6 &   6 \\
P9  &   5 &   5 &   5 &   5 &   2 &   3 &   5 \\
P10 &   7 &   7 &   7 &   7 &   6 &   5 &   7 \\
P11 &   6 &   6 &   4 &   5 &   5 &   4 &   4 \\
P12 &   4 &   6 &   6 &   6 &   3 &   2 &   2 \\
P13 &   6 &   6 &   7 &   5 &   5 &   3 &   3 \\
P14 &   1 &   6 &   5 &   5 &   3 &   2 &   2 \\
P15 &   6 &   7 &   7 &   7 &   6 &   6 &   5 \\
\bottomrule
\end{tabular}
\end{table}

\begin{table}[H]
\centering
\caption{Individual Wearables Likert Raw Data}
\vspace{-0.4cm}
\label{tab:individual_W}
\begin{tabular}{lccccccc}
\toprule
\textbf{Participant} & \textbf{L1} & \textbf{L2} & \textbf{L3} & \textbf{L4} & \textbf{L5} & \textbf{L6} & \textbf{L7} \\
\midrule
P1  &   3 &   6 &   5 &   5 &   4 &   5 &   3 \\
P2  &   7 &   7 &   7 &   7 &   6 &   6 &   6 \\
P3  &   6 &   6 &   6 &   6 &   6 &   6 &   6 \\
P4  &   3 &   7 &   6 &   7 &   6 &   5 &   7 \\
P5  &   3 &   3 &   3 &   3 &   2 &   5 &   3 \\
P6  &   7 &   7 &   7 &   7 &   3 &   2 &   2 \\
P7  &   3 &   7 &   6 &   6 &   6 &   6 &   5 \\
P8  &   6 &   6 &   6 &   6 &   5 &   6 &   6 \\
P9  &   5 &   5 &   5 &   5 &   2 &   3 &   5 \\
P10 &   7 &   7 &   7 &   7 &   6 &   5 &   7 \\
P11 &   6 &   6 &   4 &   5 &   5 &   4 &   4 \\
P12 &   4 &   6 &   6 &   6 &   3 &   2 &   2 \\
P13 &   6 &   6 &   7 &   5 &   5 &   3 &   3 \\
P14 &   1 &   6 &   5 &   5 &   3 &   2 &   2 \\
P15 &   6 &   7 &   7 &   7 &   6 &   6 &   5 \\
I1 & 4 & 7 & 6 & 6 & 6 & 5 & 7 \\
I2 & 6 & 7 & 7 & 7 & 6 & 6 & 6 \\
\bottomrule
\end{tabular}
\end{table}

\begin{table}[H]
\centering
\caption{Social Fixed Interval Likert Raw Data}
\vspace{-0.4cm}
\label{tab:social_FI}
\begin{tabular}{lcccccccc}
\toprule
\textbf{Participant} & \textbf{L1} & \textbf{L2} & \textbf{L3} & \textbf{L4} & \textbf{L5} & \textbf{L6} & \textbf{L7} & \textbf{L8}\\
\midrule
P1  &   6 &   3 &   2 &   2 &   4 &   5 &   5 &   2 \\
P2  &   6 &   6 &   6 &   6 &   6 &   7 &   7 &   7 \\
P3  &   4 &   2 &   1 &   5 &   3 &   6 &   6 &   5 \\
P4  &   7 &   4 &   5 &   7 &   6 &   7 &   7 &   6 \\
P5  &   6 &   3 &   5 &   6 &   6 &   7 &   7 &   6 \\
P6  &   1 &   1 &   1 &   1 &   1 &   1 &   7 &   2 \\
P7  &   4 &   1 &   2 &   3 &   3 &   6 &   6 &   4 \\
P8  &   2 &   2 &   2 &   2 &   2 &   3 &   6 &   3 \\
P9  &   6 &   5 &   3 &   5 &   5 &   6 &   6 &   6 \\
P10 &   3 &   2 &   5 &   7 &   5 &   2 &   7 &   6 \\
P11 &   6 &   3 &   4 &   5 &   5 &   6 &   5 &   6 \\
P12 &   6 &   5 &   5 &   6 &   5 &   5 &   5 &   5 \\
P13 &   5 &   5 &   4 &   5 &   3 &   6 &   6 &   6 \\
P14 &   3 &   5 &   3 &   5 &   5 &   3 &   7 &   5 \\
P15 &   6 &   3 &   3 &   3 &   5 &   6 &   6 &   6 \\
\bottomrule
\end{tabular}
\end{table}

\begin{table}[H]
\centering
\caption{Social Mouth Open Likert Raw Data}
\vspace{-0.4cm}
\label{tab:social_MO}
\begin{tabular}{lcccccccc}
\toprule
\textbf{Participant} & \textbf{L1} & \textbf{L2} & \textbf{L3} & \textbf{L4} & \textbf{L5} & \textbf{L6} & \textbf{L7} & \textbf{L8}\\
\midrule
P1  &   1 &   3 &   6 &   2 &   3 &   3 &   2 &   2 \\
P2  &   3 &   6 &   6 &   5 &   2 &   3 &   5 &   3 \\
P3  &   2 &   7 &   7 &   2 &   2 &   5 &   2 &   2 \\
P4  &   2 &   5 &   6 &   6 &   5 &   3 &   5 &   5 \\
P5  &   3 &   5 &   5 &   6 &   2 &   3 &   5 &   4 \\
P6  &   1 &   5 &   5 &   5 &   5 &   4 &   2 &   2 \\
P7  &   3 &   6 &   6 &   6 &   6 &   6 &   6 &   6 \\
P8  &   5 &   5 &   2 &   5 &   3 &   3 &   3 &   5 \\
P9  &   1 &   5 &   3 &   4 &   1 &   3 &   3 &   1 \\
P10 &   2 &   3 &   5 &   6 &   2 &   2 &   6 &   5 \\
P11 &   5 &   4 &   4 &   4 &   3 &   5 &   4 &   5 \\
P12 &   2 &   6 &   5 &   4 &   2 &   2 &   2 &   4 \\
P13 &   1 &   6 &   6 &   5 &   2 &   2 &   2 &   1 \\
P14 &   1 &   5 &   6 &   6 &   5 &   2 &   2 &   3 \\
P15 &   5 &   6 &   7 &   6 &   6 &   5 &   5 &   3 \\
\bottomrule
\end{tabular}
\end{table}

\begin{table}[H]
\centering
\caption{Social Wearables Likert Raw Data}
\vspace{-0.4cm}
\label{tab:social_W}
\begin{tabular}{lcccccccc}
\toprule
\textbf{Participant} & \textbf{L1} & \textbf{L2} & \textbf{L3} & \textbf{L4} & \textbf{L5} & \textbf{L6} & \textbf{L7} & \textbf{L8}\\
\midrule
P1  &   6 &   6 &   6 &   5 &   6 &   6 &   6 &   6 \\
P2  &   7 &   7 &   7 &   7 &   7 &   7 &   7 &   7 \\
P3  &   5 &   5 &   5 &   5 &   4 &   3 &   5 &   4 \\
P4  &   6 &   5 &   7 &   6 &   6 &   5 &   6 &   7 \\
P5  &   5 &   6 &   5 &   5 &   6 &   4 &   5 &   6 \\
P6  &   7 &   6 &   7 &   7 &   7 &   7 &   7 &   7 \\
P7  &   5 &   6 &   6 &   6 &   6 &   6 &   7 &   7 \\
P8  &   1 &   2 &   5 &   3 &   3 &   2 &   3 &   1 \\
P9  &   5 &   5 &   6 &   6 &   6 &   6 &   7 &   7 \\
P10 &   7 &   4 &   3 &   7 &   7 &   6 &   7 &   7 \\
P11 &   2 &   3 &   3 &   2 &   2 &   2 &   6 &   5 \\
P12 &   5 &   4 &   5 &   4 &   4 &   5 &   5 &   6 \\
P13 &   5 &   6 &   6 &   6 &   5 &   7 &   7 &   6 \\
P14 &   3 &   5 &   6 &   5 &   5 &   7 &   7 &   6 \\
P15 &   6 &   7 &   7 &   7 &   7 &   7 &   7 &   7 \\
I1 & 7 & 6 & 7 & 4 & 7 & 7 & 7 & 7 \\
I2 & 5 & 5 & 5 & 6 & 5 & 6 & 6 & 6 \\
\bottomrule
\end{tabular}
\end{table}

\subsection{Phase 2 Qualitative Responses}
\label{appendix:qualitative_responses}
We include all qualitative responses from all participants from the Phase 2 study below. 

\subsubsection{Why did you prefer your highest ranked method? Please elaborate and compare the methods for both the individual and social settings.}
\begin{itemize}
\item \textbf{P1:}``It was the most seamless and recreated the way I usually eat by waiting for me to stop chewing and moving before bringing the spoon closer to my mouth. In the social setting, it waited until I stopped talking before bringing the spoon closer and if I was in the middle of a conversation, I could easily stop it from coming closer. In the individual setting, it just followed a good pace for eating.''
\item \textbf{P2:} ``I preferred the highest ranking method because it required the least amount of thinking of like should i go for a bite or should i not in a social setting especially. I also liked it the most in the individual setting because it would wait for me to finish chewing or be ready. I did not have to think about if I should signal for a bite or if I need to chew faster.'' 
\item \textbf{P3:} ``I think the wearable method was especially beneficial for the social setting, since the actions for pausing the feeding aligned well [with] the social cues where I'd rather not be disrupted. For the individual setting, I think the awareness of chewing served a similar function. Assuming I'd like to be fed continuously, the wearables introduced brakes for situations where I need a bit more time.'' 
\item \textbf{P4:} ``Within the social setting, the timing of each bite was appropriate, never interfering with the conversation. Within the individual setting, it allowed me to eat at a faster pace, also giving the bite at an appropriate time.''
\item \textbf{P5:} ``During the individual portion, the wearables method was able to adjust and bring food faster than the fixed portion one. In the social setting the wearables and fixed interval were both similar, since I wasn't eating as quickly as the individual portion. The mouth open method was a bit distracting, and the detection for when my mouth was open was a bit inconsistent. I had to keep my mouth very wide open for a pretty long time for it to work.''
\item \textbf{P6:} ``I thought it allowed me to have a seamless conversation without needed to do any extra or unnatural movements. It was also a seamless individual experience because I was not producing any head or throat movements, and therefore the robot did not stop on the way to my mouth.''
\item \textbf{P7:} ``The wearables had the best timing for solo dining since I did not have to wait between bites, and it did not deliver the food too early. For social dining, this method also had the best timing but was slightly distracting for the conversation, though still better than the other methods.''
\item \textbf{P8:} ``I liked that the fixed interval was the most predictable. I could more easily plan my conversation topics when I knew the robot was on a schedule. For individual eating, I thought this method was on the slow side, but I think I still preferred this method. For social interactions, this method was preferable for the reason stated above, namely predictability.''
\item \textbf{P9:} ``I prefer the wearables approach since it did not require me to actively do anything (compared to mouth open) or to needlessly wait (as in fixed interval). It was particularly helpful in the social context where it did not require me interrupting the conversation.
\item \textbf{P10:} ``The wearables (my highest ranked) method was very intelligent and felt seamless in both scenarios.''
\item \textbf{P11:} ``when i'm on my own, fixed interval seems slow and ineffective, even annoyingly suspenseful. In a conversation, it's nice to have time to chew, drink, talk and not worry about the robot. the other methods that asked me to act in order to get food or stop the food from coming got in the way of conversation and were additionally cumbersome because the robot couldn't accurately react to my signals for giving me/stopping the food. the robot couldn't see when i opened my mouth asking for food. i'm sure this would be much, much worse for individuals who are farther from the norm in computer vision datasets. the same is true with the wearables signals stopping the food from coming. i wanted food to come but i was also talking. i can talk with food in my mouth. this isn't 1700s france.''
\item \textbf{P12:} ``The wearables method allowed for the best overall timing of the bites. In the social setting, it was pretty seamless and I was pretty distracted by the conversation to notice the robot waiting or needing any direction from me to proceed with its trajectory. For the individual setting, I thought it moved slightly quickly but overall timed the bites well without making me wait too long.''
\item \textbf{P13:} ``The wearables method had the least mental workload for me, and I felt like I didn't have to do much on my end or wait significantly long for the food to reach my mouth. For the individual portion, it felt most like I was having a natural meal and the timing felt most appropriate, and likewise for the social dining portion. The fixed interval portion was not distracting, but the timing for food often felt prolonged or rushed depending on the rate I was consuming the food. For the mouth open portion, the robot's commands felt distracting, especially in a social setting where I felt obligated to open my mouth in the middle of conversation.''
\item \textbf{P14:} ``I liked the wearables method because it required the least amount of thought during the social and individual setting. It felt the most seamless and like it didn't interrupt my train of though while eating.''
\item \textbf{P15:} ``It required the least amount of work on my part and had the best timing. I didn't have to open my mouth or wait unnecessarily long for the food to come. Additionally, in the social setting I could continue having a conversation and the robot would stay still and wait. Thus, it wouldn't make the conversation awkward. I felt like mouth open made the conversation a bit awkward. Additionally, because I was told to take the bite as soon as possible for fixed interval it also made conversation a bit awkward.''
\end{itemize}

\subsubsection{Which methods did you feel in control while using? Why/why not?}
\begin{itemize}
\item \textbf{P1:} ``I felt more in control using the wearables method because I could stop it at any time. I also felt in control for the mouth open method, but not as much in control. Once you signaled to the robot you were ready, you couldn't really change your mind about the bite, which became awkward in the social setting.''
\item \textbf{P2:} ``I felt in control using the mouth open and the wearables. In those, either it waited for me to clearly be done with my bite including chewing or it would wait for me to signal that im ready to eat.''
\item \textbf{P3:} ``I felt most in control with the mouth open condition, since it relied on an explicit cue to trigger the feeding. The wearables came second in this regard. Sometimes thought, I don't know if the actions fully captured my intent to stop the feeding. I have no control over the fixed interval setting, since that operated on a fixed cycle regardless of what I was doing.''
\item \textbf{P4:} ``I felt in control with the two non-timed methods, because they reacted to my prompting. I didn't feel as in control with the timed method, because if I wanted it sooner or it came too soon, I still had to take the bite. ''
\item \textbf{P5:} ``The mouth open made me feel most `in control' in the sense that I can most easily tell it that I didn't want to eat, because if I didn't want food in the wearables method I needed to come up with something that will pause the arm from coming in like talking or chewing (or pretending to chew). The timing one didn't have any feedback so I didn't really feel in control, but the interval was long enough that I never felt like I was rushed to eat, even if I couldn't tell it when I was ready.''
\item \textbf{P6:} ``The fixed interval did not allow me to be in control at all since it was based on a fixed time interval and not my own movements. The mouth open made me feel somewhat in control, but when it did not recognize my mouth movements, especially in the social setting, it was a bit awkward and I felt that I needed to put a lot of mental and physical effort. The wearables made me feel in control in the sense that it understood my natural body movements and I did not have to exert any additional effort in order for it to feed me.''
\item \textbf{P7:} ``The wearables and the mouth open methods since I could determine when I could receive the next bite. Mouth open had the most control but was slightly inconvenient since it interrupts the conversation.''
\item \textbf{P8:} ``I felt most in control of the mouth opening method, though there is nuance in this answer. I think the mouth opening gave me the most control over the timing of the robot, but the fixed-interval gave me the most control of social interactions. ''
\item \textbf{P9:} ``For the individual setting, I think the mouth open and wearables felt in control. However, moving my head and opening my mouth felt a bit artificial. In the social context, the wearables and fixed interval felt more controllable as the open mouth procedure was interrupting the whole dining flow.''
\item \textbf{P10:} ``I felt the most in control with the mouth open method because I could signal to the camera whenever I wanted to eat, although I had to keep my mouth open for a while. The other two methods gave me less sense of control because I didn't give an explicit signal for being ready to eat.''
\item \textbf{P11:} ``ironically, fixed interval. wearables was pretty annoying as i had to stay unnaturally still to get food. mouth open just didn't detect my mouth a lot of the time.''
\item \textbf{P12:} ``I felt in control using the mouth open method, since I could tell the robot to give me the next bite pretty explicitly by opening my mouth. I thought the wearables sometimes made me feel control, but the sensors sometimes didn't pick up on the fact that I was still chewing. The fixed interval method didn't really look for when I was done with the bite to give the next one, so the timing was pretty late.''
\item \textbf{P13:} ``I felt most in control while using the mouth open and wearables methods because the robot actively responded to my behavior. The wearables method felt more seamless since I didn't have to perform the action of opening my mouth for every bite. For the fixed interval, I felt like the robot was in more in control of when I took bites, since I couldn't really do anything to signal that I wanted to take bites or wanted to take a break.'' 
\item \textbf{P14:} ``I felt in control when using the wearables and the mouth open because it either looked or waited for some sort of cue from me. The fixed interval one was just based on time, so it didn't feel like I could control any part of it.
\item \textbf{P15:} ``Mouth open. The robot wouldn't move until you gave it the command to.''
\end{itemize}

\subsubsection{Which method had the most appropriate bite timing? Elaborate on why it had the most appropriate timing and compare it to the other methods.}
\begin{itemize}
\item \textbf{P1:} ``Definitely the wearables one. It was the easiest to use and I could stop it / let it continue whenever I wanted. The fixed interval one was okay during the social setting because I would just take a break from my conversation whenever the robot brought the spoon closer to my mouth, but there was a lot of waiting when there was no social context. The mouth open was the most awkward during the social setting because you had to manually signal to the robot during the conversation. It was better in no social context because you could time the bites better.''
\item \textbf{P2:} ``The wearables had the most appropriate bite timing because compared to the other methods, it would wait for me if I started chewing or talking after I signaled that I was ready to eat. I didn't have to control the conversation for a pause around the time that the bite would be ready. In terms of the individual portion, I liked this timing the best because it didn't force me to wait for another bite like the other ones, it kind of naturally matches my pace, the other ones felt a little slow or fast at times.''
\item \textbf{P3:} ``I think the wearable provided the most appropriate bite timing. I think they were all not horrible though, including the fixed interval. Strangely, the fixed interval felt most appropriate during the social setting since it was easier to anticipate.''
\item \textbf{P4:} ``Within the social context, I'd say both fixed interval and wearables had the most appropriate timing. 45 seconds felt too long when eating individually, but worked well within the conversation. The wearables also knew when I was ready to take a bite, both within the context of a conversation and individually. The mouth open method had a delay between when I wanted the bite and when I was given the bite, and it was awkward at times to turn my head, or to open my mouth in the middle of a conversation.'' 
\item \textbf{P5:} ``The wearables one felt the most appropriate across both portions, since it would slow down the feeding interval during the conversation portion, but was able to adjust to the individual portion when I wanted bites more frequently. The mouth open one has good timing too since I could tell it exactly when to bring the next bite, but since it took a while for the robot to detect my open mouth, the timing felt a little slow.''
\item \textbf{P6:} ``The wearables method had the most appropriate bite timing because it was based on my natural body movements. Sometimes the mouth open method did not recognize my mouth opening, so I had to open wider or open more times, which took up a lot of effort and time. The fixed interval method was not convenient in terms of the social setting because it couldn't understand my social cues/behavior and make the bite timing more appropriate based on that.''
\item \textbf{P7:} ``The wearables had the best timing for individual dining since it delivered a bite whenever I was nearly done chewing. For social dining, the wearables was slightly distracting since sometimes during the conversation, I wanted to wait until the other person stopped talking before taking a bite (even if I wasn't moving, chewing, or talking).''
\item \textbf{P8:} ``In social dynamics I think timed eating was the best option for me. The wearable method did not offer me much of a benefit. It seemed too quick for the pace of conversation. Because I speak more slowly and take time to think between ideas, I felt the robot was smothering me with food when I was hoping to talk.''
\item \textbf{P9:} ``The wearables definitely had better bite timing as I did not need to needlessly wait or artificially open my mouth.''
\item \textbf{P10:} ``The wearables method was the best. I really liked how the robot spoon would start approaching me slowly as I was getting more ready to eat (the rate of movement also felt more appropriate and adaptive depending on my rate of chewing), as opposed to the fixed interval method which would move quite fast.''
\item \textbf{P11:} ``fixed interval. there was simply too much effort and time and focus dedicated to controlling the robot. i can just take the bite when it comes if i have enough time in between to chew and drink and talk. fixed interval enabled that. I spent most of the time trying to communicate with the robot in the wearables and open mouth method.''
\item \textbf{P12:} ``Overall the mouth open method had the most appropriate bite timing, since I signaled when I wanted the next bite. The wearables method was overall slightly early, as the sensors didn't always pick up when I was still chewing. The fixed interval method had good timing for the social setting but was late for the individual setting, so I was kind of just sitting there waiting for the next bite.''
\item \textbf{P13:} ``I think the mouth open method had the most appropriate bite timing because the robot only moved when I signaled for it to, and I could take things at my own pace (i.e. taking a break from eating if I wanted to). The wearables method had the second most appropriate bite timing since the robot moved according to the pace I was eating and talking. The fixed interval method, as mentioned, often felt rushed or prolonged for timing between bites.''
\item \textbf{P14:} ``I think the fixed interval one was good during the social setting because it gave me a substantial amount of time during each bite. I didn't mind it during the individual one too because it allowed me to be mindful of my bites. I think the most appropriate timing was the wearables though because it was able to understand when I was finished chewing.''
\item \textbf{P15:} ``Wearables. It would wait for me while I was chewing and talking. Compared to mouth open, the robot would remind me to open my mouth and wouldn't come until I did. This meant that I had to pause conversation, turn towards the robot, open my mouth, and wait for it to come which disrupted conversation a bit. Additionally, fixed interval seemed to make the robot come at random points during the conversation which also disrupted conversation a bit.''

\end{itemize}

\subsubsection{What cues do you think a caregiver/you use to know when to feed a care recipient/yourself a bite?}
\begin{itemize}
\item \textbf{P1:} ``stop chewing and stop talking''
\item \textbf{P2:} ``When i am still for a bit, I think its a good indication of when to feed me a bite, or more specifically if I am still for a bit and maybe look at the food for a bit too.'' 
\item \textbf{P3:} ``While I did not appreciate the experience, I think probably some sort of explicit gesture or cue would easiest.''
\item \textbf{P4:} ``Finishing a swallow, not talking, expectant look, opening mouth. ''
\item \textbf{P5:} ``Looking at the food, stopping conversation''
\item \textbf{P6:} ``Head turning or mouth opening''
\item \textbf{P7:} ``Caregiver: A minor physical cue without the movement of the head, like a finger tap. Myself: When eating alone, I would take bites once done chewing. For social dining, I would take a bite in natural pauses in the conversation.''
\item \textbf{P8:} ``I am curious if a nod would be helpful. The open mouth seems like it could have worked, though the camera seemed to see me speaking as cuing the robot.''
\item \textbf{P9:} ``Mainly determining if the person's mouth who is eating is ready or not; i.e., not empty. In a social context, it would be knowing a pause would occur where the person could grab a bite.''
\item \textbf{P10:} ``The amount of talking, the speed of chewing''
\item \textbf{P11:} ``when i fed my sick grandmother, we would talk, and i would wait until  she was done chewing and drinking to give another bite. this is also how we do with babies. robots simply don't have that amount of sensitivity and our academic and research institutions aren't nearly as equitable as they'd need to be to build models that detect this non-verbal cues in people with different backgrounds, needs, abilities, etc. if i'm feeding someone who is farily sick, i would ask. voice recognition may be something. but then again, many automated voice systems are discriminatory to accents and various dialects not reflecting the elite.''
\item \textbf{P12:} ``I think a caregiver would probably look for when a recipient has swallowed and looks like they have finished chewing their bite to give them the next one.''
\item \textbf{P13:} ``I think the cues that a caregiver would use to know when to feed a care recipient a bite would include some sort of communicated signal (mouth opening or a nod) or simply seeing when the recipient had stopped chewing and swallowing.''
\item \textbf{P14:} ``The cues I use are when I am done chewing, when I am done talking (social setting). For others (recipients), I think I would wait for a head nod or some physical/visible signal.''
\item \textbf{P15:} ``When the recipient isn't chewing or talking. This indicates that the recipient is done taking the previous bite and that they're available for the next bite.''
\end{itemize}

\subsubsection{Which method was closest to how a caregiver/you would perform this task? Please elaborate on why.}
\begin{itemize}
\item \textbf{P1:} ``Definitely the wearables one.''
\item \textbf{P2:} ``I think the wearables is the closest to how I would perform the task because, when I change my mind and want to take a little longer between signaling for a bite and actually eating it, it matches my time so i dont really feel rushed or I dont feel frustrated with the slowness.''
\item \textbf{P3:} ``Probably the mouth open or the wearable. The former is more like asking for each bite, while the later is effectively assuming the recipient would like to be continuously fed and waiting for somewhat appropriate moments to initiate.''
\item \textbf{P4:} ``Mouth open, because it's responsive for certain, so you know the person wants another bite.''
\item \textbf{P5:} ``I think the wearables was the closest, since it wouldn't give the bite when I clearly didn't want it, and it would adjust to the environment to decide when to give the next bite, like a person would.''
\item \textbf{P6:} ``The mouth open method because I assume mouth opening is a cue for caregivers to give a scoop.''
\item \textbf{P7:} ``Wearables, the timing for individual dining was the best, and while it was slightly distracting during social dining, it was still the best among the three in terms of timing and interruption.''
\item \textbf{P8:} ``I think the wearable method would best simulate a caregiver in theory; however, I did not find this to be true for my experience. I would think that open-mouth cuing would be next for simulating a caregiver. I think a hybrid method could be the best option. one where an open mouth cues the motion, but the wearable gives feedback while it is en-route to slow or stop the robot if necessary.''
\item \textbf{P9:} ``The wearables was closest as it dynamically waited but it did pause a bit on its trajectory which was a bit distracting.''
\item \textbf{P10:} ``Wearables method.''
\item \textbf{P11:} ``fixed interval. i was simply given enough time to eat and drink and talk. it wasn't a chore.''
\item \textbf{P12:} ``I think the wearables method is probably closest to how I would perform this task, since it looks for the recipient for feedback as to when the next bite should be given, without forcing them to be very explicit/make it feel incredibly unnatural.''
\item \textbf{P13:} ``I think the last method (wearables) was closest to how a caregiver would perform this task since it was uninterrupted and felt like the most natural timing according to the caregiver/robot's observations of the recipient's eating patterns.''
\item \textbf{P14:} ``I think the mouth open because it is a visual cue. The timed one, I dont think a caregiver would use because it's too rigid of a method. The wearables, I think it may be difficult for a caregiver to know exactly when the person is done chewing.''
\item \textbf{P15:} ``Wearables. I would only give the recipient food when they want it and when it's natural. I think this was most successfully accomplished by the wearables. Only feeding at fixed intervals is a bit awkward since you could be disrupting the recipient's conversation/chewing. Additionally, although mouth open guarantees that the recipient wants food and is available to eat, it disrupts the conversation so it feels less natural.''
\end{itemize}

\subsubsection{Were the glasses comfortable to wear while eating? If not, please explain.}
\begin{itemize}
\item \textbf{P1:} ``I am used to wearing glasses, so the glasses where comfortable to me.''
\item \textbf{P1:} ``Yes they are comfortable, but it may be due to already wearing glasses on a normal basis.''
\item \textbf{P3:} ``Did not bother me. ''
\item \textbf{P4:} ``I usually wear glasses, so visually I didn't have any discomfort. After taking them off, I noticed that they were a bit heavy, but I hadn't noticed that strain while wearing them.''
\item \textbf{P5:} ``They were about as comfortable as normal glasses. Although the bottom brim of the glasses did block my view a little, making it harder to see the spoon.''
\item \textbf{P6:} ``They started falling down but I think that is just because I have a low nose bridge. I don't think this would be an issue for other people.''
\item \textbf{P7:} ``Yes''
\item \textbf{P8:} ``Yes, I did not notice them.''
\item \textbf{P9:} ``Yup, they felt natural.''
\item \textbf{P10:} ``The glasses were generally comfortable; they were quite lightweight but as someone who doesn't ordinarily wear glasses, the frames felt a bit thick and changed my field of view as a result. There was no physical discomfort otherwise.''
\item \textbf{P11:} ``I hardly noticed the glasses. they protected me from getting hit in the eye by the robot, in fact.''
\item \textbf{P12:} ``The glasses were comfortable to wear while eating.''
\item \textbf{P13:} ``The glasses were comfortable (since they were my own glasses)--the additional sensor was also not too distracting.''
\item \textbf{P14:} ``Yes.''
\item \textbf{P15:} ``They'd slip down my nose occasionally, but I didn't find it too much of a bother.''
\end{itemize}

\subsubsection{Was the throat microphone comfortable to wear while eating? If not, please explain.}
\begin{itemize}
\item \textbf{P1:} ``Yes, I stopped noticing it once the trial started.''
\item \textbf{P2:} ``Yes, I didn't really notice it that much as it wasn't that tight and I forgot about it quickly after I put it on.''
\item \textbf{P3:} ``Did not bother me.''
\item \textbf{P4:} ``I didn't notice it very much.'' 
\item \textbf{P5:} ``I barely noticed the microphone, probably because I wasn't moving much.''
\item \textbf{P6:} ``It was a bit uncomfortable in the beginning but I forgot it was there after awhile.''
\item \textbf{P7:} ``Yes''
\item \textbf{P8:} ``Yes, I also did not notice this''
\item \textbf{P9:} ``They were initially distracting when put on, but afterwards when listening to instructions, eating, talking in the social setting, etc I forgot I was wearing them.''
\item \textbf{P10:} ``The throat microphone was fine although I wouldn't want to wear it for a prolonged period of time due to the fact that I don't like having something touch my neck all the time.''
\item \textbf{P11:} ``it felt annoying in the beginning. but then in conversation I forgot about it. different levels of anxiety could see a device like that bring a lot of stress to someone.''
\item \textbf{P12:} ``The throat microphone was okay to wear while eating, though it was kind of tight and I could definitely feel it during the experiment.''
\item \textbf{P13:} ``The throat microphone was relatively comfortable to wear while eating, and once I got used to it I could eat without it being a distraction. It was a little uncomfortable in the beginning, and especially when I was not eating since I could feel it putting pressure on my throat.''
\item \textbf{P14:} ``Yes.''
\item \textbf{P15:} ``Yes, I didn't really feel it during the test.''
\end{itemize}

\subsection{Phase 3 Qualitative Responses}
\label{appendix:phase3_questions_responses}
The 2 individuals with motor impairments answered the following questions verbally after both the individual and social dining sessions and their responses were transcribed after. I1 did the individual session first while I2 did the social session first. 

\subsubsection{Did you like this bite timing method for this setting? Why or why not?}
\begin{itemize}
    \item \textbf{I1 (Individual):} ``I do like it. Cause I’m in control of the meal, when I was gonna take the bite, when I wasn’t. Like before, it seemed pretty … to flow pretty good. You know? So if it was just me sitting there eating dinner then it was pretty good. I liked it.''
    \item \textbf{I1 (Social):} ``Yes, because it just let me have a conversation and eat at the same time.''
    \item \textbf{I2 (Individual):} ``Yes, I like it, um… I didn't have to think about it, it was...The timing was great, it was… I loved it really.''
    \item \textbf{I2 (Social):} ``Yes... Um, I liked it because I didn't have to… stop what I was doing and do a specific action. And be like, hold on, don’t talk for a second. You know, just, um… like I said it was almost seamless, it was...I didn't really have to think about it. It's just the most I had to think about was, shutting up and not moving, so I can eat. But otherwise, yeah, I...because there was no other action involved I really liked it.'' 
\end{itemize}

\subsubsection{Did you feel in control while using the robot with this bite timing method for this setting? Why/why not?}
\begin{itemize}
    \item \textbf{I1 (Individual):} ``Yes. It stopped when I wanted to. From the commands you gave me, it seemed to work pretty good.''
    \item \textbf{I1 (Social):} ``Yes... It just seemed to go pretty good with the commands you gave it.''
    \item \textbf{I2 (Individual):} ``Um, yeah, for the most part, I think one time I...Moved my head, and it didn't listen to me, but it could have been a slighter move that I didn't realize was little or, but it wasn't a big deal, it's not. Everything else it responded to perfectly.''
    \item \textbf{I2 (Social):} ``For the most part. Um, it's a little weird feeling just this fork coming at you, you know? Um, but it was controlled enough that I wasn't afraid that it was gonna hit me in the face without me stopping first.''
\end{itemize}

\subsubsection{Did this method have appropriate bite timing for this setting? Elaborate on why/why not.}
\begin{itemize}
    \item \textbf{I1 (Individual):} ``Yeah. No it was alright. Because the robot can only go so fast...I thought it was pretty, pretty good.''
    \item \textbf{I1 (Social):} ``I think pretty much most of the time, yeah. It just seemed like when I was not chewing and it just when I turned and it was like `oh, I'll take a bite of this' they were waiting for me.''
    \item \textbf{I2 (Individual):} ``Yup... see above.''
    \item \textbf{I2 (Social):} ``Um, yeah, I would say so. Um,...I don’t know how much more to say. But yeah, I like this setting. the setting, the bite timing, I... it just seemed to work.''
\end{itemize}

\subsubsection{What cues do you think a caregiver uses to know when to feed you a bite for this setting?}
\begin{itemize}
    \item \textbf{I1 (Individual):} ``If it was a caregiver. Probably watching when I was done chewing. Like, if I turn toward them, obviously, they’d think you know it was time to give me a bite.''
    \item \textbf{I1 (Social):} ``Um, probably whether I’m still chewing or not. Different movements or mannerisms. Yeah, if there's a pause in the conversation maybe... or if I just stop talking. Maybe they're like `oh, he wants a bite'.''
    \item \textbf{I2 (Individual):} ``If we’re watching TV, forget about it. This [the robot] is so much better, cause if we’re watching TV, they don’t watch me as much as they watch the TV. You know, a lot of times. I mean some are better than others, but yeah. That’s.. They don’t pay attention that’s... So yeah this is great.'' 
    \item \textbf{I2 (Social):} ``I've actually told some of them `watch for me chewing.' You know, if I’m done.. if I'm still chewing, you don't need to pick the fork up. You don't, you know, as far as that, you don't need to...come toward me with it. I'll tell you a funny story real quick. One of my caregivers kept putting the fork so close, you know, like, here, have a bite, and I still had a mouthful. I said, next time you do it, I'm gonna show you what's in my mouth. And I did. And I opened my mouth, and I still had, you know, because if a bite's big, it takes you longer to chew. Um, but yeah, so that's really what I’ve told them to look for, too. Mostly that, and...It drives me crazy when I'm talking to somebody, and they're feeding me, and they go to give me a bite. You know, so I have to tell them when the other person's talking is when you need to give me the bite.''

\end{itemize}

\subsubsection{Was this method close to how a caregiver would perform this task for this setting? Please elaborate on why.}
\begin{itemize}
    \item \textbf{I1 (Individual): } ``I would say yes. Because when I wanted a bite, I got it... it was just a different pathway I guess. Rather than me asking for a bite or turning, opening my mouth and doing basically the same thing with a robot but just keeping my head still and not chewing or talking....Different chain of commands.''
    \item \textbf{I1 (Social):} ``I would say yes. It didn’t seem to take away from the conversation, so it was just kind of part of that environment I guess?...It just seemed to not interrupt the conversation.''
    \item \textbf{I2 (Individual):} ``No, way better, way better than they would do it...Way better just to be truthful. You hear that?''
    \item \textbf{I2 (Social):} ``Oh, I kind of answered that, right? So, yeah. It waited for me to stop moving and talking and chewing. Yeah.''
\end{itemize}

\subsubsection{Were the glasses comfortable to wear while eating for this setting? If not, please explain.}
\begin{itemize}
    \item \textbf{I1 (Individual):} ``Yeah they were ok.'' 
    \item \textbf{I1 (Social): } ``Yeah, they weren’t too bad.''
    \item \textbf{I2 (Individual):} ``Oh, there we go, yeah. I didn’t even notice them.''
    \item \textbf{I2 (Social):} ``Yeah, actually. When you... well, when you first put this thing on, I'm like, I got this piece of tape over there, but I forgot it was there after a while... The black tape probably would’ve… wouldn’t have been as obvious, but yeah, once we started eating, I forgot it was there.''
\end{itemize}

\subsubsection{Was the throat microphone comfortable to wear while eating for this setting? If not, please explain.}
\begin{itemize}
    \item \textbf{I1 (Individual):} ``I got used to it pretty quick.'' 
    \item \textbf{I1 (Social):} ``Yeah, it was fine. ''
    \item \textbf{I2 (Individual): } ``That's the same thing, too, is it's... not used to it on my neck, but it’s not so uncomfortable that I'm like, I can't do it. You know, it's just...it's...what's the word? Awkward, but not uncomfortable.''
    \item \textbf{I2 (Social):} ``Um, it's not bad when you first put it on, I was like, Because I'm not a big like turtle neck, tight neck person, but I kind of forgot about that, too. So it's a little snug, but it's not awful.''
\end{itemize}

\end{document}